\documentclass{article}

\usepackage{PRIMEarxiv}

\usepackage[utf8]{inputenc} 
\usepackage[T1]{fontenc}    
\usepackage{hyperref}       
\usepackage{url}            
\usepackage{booktabs}       
\usepackage{amsfonts}       
\usepackage{nicefrac}       
\usepackage{microtype}      
\usepackage{lipsum}
\usepackage{fancyhdr}       
\usepackage{graphicx}       
\graphicspath{{media/}}     

\usepackage{amssymb}
\usepackage{amsthm,amsmath}
\usepackage{lscape}
\usepackage{xcolor}
\usepackage{soul}
\usepackage[skip=2pt]{caption,subcaption}
\usepackage{tabularray}
\usepackage{comment}
\usepackage{algorithm,setspace}
\usepackage{algpseudocode}
\title{Adaptive operator selection utilising generalised experience
}

\author{
  Mehmet Emin Aydin\\
  School of Computing and Creative Technologies \\
  UWE Bristol \\
  Bristol, UK\\
  \texttt{mehmet.aydin@uwe.ac.uk} \\
    \And
  Rafet Durgut \\
  Department Computer Engineering\\
  Bandirma Onyedi Eylul University \\
  Bandirma, Turkiye\\
  \texttt{rdurgut@bandirma.edu.tr} \\
 \AND
  Abdur Rakib \\
  Institute for Future Transport and Cities (IFTC)\\
  Coventry University \\
  Coventry, UK \\
  \texttt{ad9812@coventry.ac.uk} \\
}

\begin{document}
\maketitle

\begin{abstract}
Optimisation problems, particularly combinatorial optimisation problems, are difficult to solve due to their complexity and hardness. Such problems have been successfully solved by evolutionary and swarm intelligence algorithms, especially in binary format. However, the approximation may suffer due to the the issues in balance between exploration and exploitation activities (EvE), which remain as the major challenge in this context. Although the complementary usage of multiple operators is becoming more popular for managing EvE with adaptive operator selection schemes, a bespoke adaptive selection system is still an important topic in research. Reinforcement Learning (RL) has recently been proposed as a way to customise and shape up a highly effective adaptive selection system. However, it is still challenging to handle the problem in terms of scalability. This paper proposes and assesses a RL-based novel approach to help develop a generalised framework for gaining, processing, and utilising the experiences for both the immediate and future use. The experimental results support the proposed approach with a certain level of success. 
\end{abstract}

\keywords{Operator selection \and Reinforcement Learning \and Transfer learning \and General problem solver}

\section{Introduction}
\label{sec:intro}
One of the fundamental issues with the optimisation problems is how to maintain a balance between the exploration and exploitation during the search process. One possible solution to this issue is the multiple operator neighbourhood structure, which enforces the usage of different operators subject to circumstantial suitability. The search process is a series of decision-making phases, where an operator is selected to apply a move operation to the solution in hand. Operator ordering or online selection both require a set of rules to be followed. In the literature, popular approaches include probability matching~\cite{chen2019self}, adaptive pursuit~\cite{belluz2015}, and variations of multi-armed bandit approaches~\cite{thierens2005adaptive}. These approaches have been successfully used in numerous research~\cite{li2013adaptive, scott2010modern}. These approaches, however, require applying a specific rule for all circumstances, which typically ignores dynamically changing search conditions and does not result in behaviours for unforeseen changes and dynamic situations. Machine learning and data-driven approaches are well known for addressing dynamically changing circumstances. 

Reinforcement learning has recently been used to train online agents so they can learn when and how to select each operator depending on the problem state~\cite{durgut2021adaptive_1}. However, the problem state was represented as a binary set with a size equal to the cardinality of the problem. This imposes restrictions on working with problem instances in the same size, not larger or smaller. The emerging issue is that how the agents trained with the data collected from problems in a certain size can handle the problems in different sizes, which appears as scalability issue with the approach. The problem state, on the other hand, does not include any information related to the fitness landscape and the circumstances in the neighbourhood. An approach for combining state information from the problem state and the surrounding area in order to more accurately predict the behaviour of each alternative operator. The dynamic changing properties of the search process make the learning and mapping the states to actions more difficult, which is another outstanding issue. For instance, the probability of moving to a neighbouring state in better quality is much easier in the earlier stages pf search space, while it is much harder in the very last stage. The search space can be more conveniently represented so that the nuances can be used to distinguish between the circumstances. Once the process has been successfully managed in these regards, the gained experience need to be generalised in order for the agents to transfer their knowledge and expertise to handle and solve other similar problems.     

The following contributions have been made in this paper: (i) multi-operator neighbourhood functions orchestrated with reinforcement learning; (ii) feature-based input space is used to identify/characterise and represent the problem states; (iii) search space is re-defined subject to the stages of the optimisation process; and (iv) learning/trained agents are enabled to transfer their experiences across a number of problem instances without size restriction.

The rest of the paper is organised as follows. Section~\ref{sec:lit-rev} introduces the basic underlying concepts alongside an overview of the related works, Section~\ref{sec:RL-OS} describes the proposed approach, Section~\ref{sec:expr-results} provides the details of the experimental results along with relevant discussions and analysis, and finally Section~\ref{concs} concludes the paper by critically evaluating the results and outlining directions for the further work. 

\section{Background and Literature review}
\label{sec:lit-rev}
The proposed study combines a number of approaches to design an adaptable transfer learning search strategy that takes into account multi-operator neighbourhood functions orchestrated with reinforcement learning. In order to keep the article self-contained, this section provides an overview of the necessary background information before moving on to a review relevant to the proposed work. Operator selection, which is the selection of the most suitable operator for a specific search problem, is crucial in optimisation problems because the problem (instance) being solved has a significant impact on the algorithm's performance and the correctness of the result. However, when utilising machine learning approach for operator selection, the performance of the operator selection model depends on the data used to train and test the model. Since operator selection and transfer learning are the subject of this study, here is where our review and discussion will be focused.

\subsection{Artificial Bee Colony}
\label{sec:abc}
Artificial bee colony (ABC) is the swarm intelligence algorithm used in this study as the problem solving framework ~\cite{karaboga2009comparative}. It is rather recent and widely used algorithm that turned to be success proven in solving many real \cite{dugenci2015creep} and artificial problems \cite{agarwal2016systematic}. The framework is setup inspiring of natural collectivism (collective behaviour) in honey bees in which a population of solutions is adopted to imitate a bee colony, where different types of bees are retreated in different ways so as to keep the population diverse and efficiently approximating. The structure of the algorithm does advantageously not constrain with much compulsory rules, which must be followed. Due to that algorithmic properties, ABC facilitates use of multiple operators with very high capacity. Pseudo code for the ABC algorithm implemented in this study is provided in the appendix with Algorithm \ref{alg:ABC}. 

\subsection{Reinforcement Learning}
\label{sec:RL}
Reinforcement Learning (RL) is the process of discovering what to perform by an agent in order to maximise a numerical reward signal~\cite{Sutton1998}. In the recent past, RL has become a widely used machine learning technique to formulate numerous engineering and scientific problems. The four fundamental components of the RL process are an action, a state, a reward, and an environment. In order to receive a reward, an RL agent acts at an appropriate time in an environment. The RL agent, however, is initially unaware of whether the action will have a positive or negative reward. Hence, the agent runs an exploration process to examine all possible states and behaviours. The agent receives reward as a result of each action performed in the preceding step. The agent evaluates the reward and stores it for later use. The agent can also exploit its knowledge and choose an action with the highest expected reward. The learning process is continued until it finds an optimal policy. Designing reinforcement learning systems presents a significant challenge in balancing the trade-off between exploration and exploitation.

\subsection{Operator Selection Problem}
\label{sec:os}

\begin{figure}[h!]
    \centering
    \includegraphics[width=.8\textwidth]{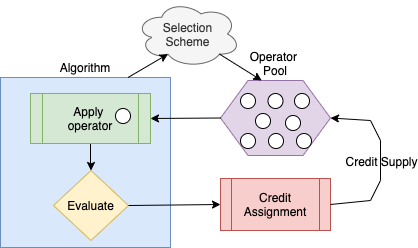}
    \caption{The overall information flow through the algorithm and other components including operator selection process}
    \label{fig:system-model}
\end{figure}

The architecture of the underlying system and the information flow are shown in Figure~\ref{fig:system-model}. As stated before, the goal of this study is to establish a comprehensive framework that will enable optimisation algorithms to employ immediate, temporary, and generalised experience while using population-based heuristic optimisation techniques, particularly to address combinatorial problems. The optimisation algorithm considered in this study---which will be referred to as~\textit{Algorithm} henceforth---is a standard Artificial Bee Colony (ABC) algorithm~\cite{karaboga2009comparative}, which is one of recently developed successfully tested swarm intelligence algorithms using population of solutions. The proposed~\textit{Algorithm} has two important components:~\textit{Apply operator} and \textit{Evaluate}, which apply a selected operator to the chosen solutions and then evaluate with respect to the fitness with the objectives.

Let $\mathcal{O}$ be a finite set of operators and $\mathcal{X}$ be the solution space. The Algorithm $\mathbf{A}$ is defined as $\mathbf{A}=\{A_t\mid t=0,1,\ldots,T\}$, a process of $T$ number of iterations---each represented with $A_t$, which uses $\langle o_i,\textbf{x}_j,F\rangle$ as a tuple of operator $o_i\in\mathcal{O}$, solution $\textbf{x}_j\in\mathcal{X}$, and the function $F:\mathcal{X}\rightarrow\mathbb{R}$ is defined as $F(\textbf{x}_j)=\|\textbf{x}_j - o_i(\textbf{x}_j)\|$, where $o_i(\textbf{x}_j) = \textbf{x}_{j+1}$. Here, $F(\textbf{x}_j)$ measures the quality of the solution of $\textbf{x}_j$, which is also known as a fitness function. The other two key components of $\mathbf{A}$ are \textit{Credit Assignment} and \textit{Operator Pool} with which each operator $o_i\in\mathcal{O}$ to be used and evaluated, a credit $c_i\in\mathcal{C}$ is assigned to them for the usage in the future. Where  $\mathcal{C}=\{c_i|i=1, 2, \ldots, |\mathcal{O}|\}$, $c_i=\sum_{t=0}^{\pi}c_{i,t}$,  $\pi$ is the current iteration, and $c_{i,t}$ is awarded based on $F(\textbf{x}_j)$. A \textit{Selection Scheme} is a function $\sigma : \mathcal{X} \times \mathbb{R} \rightarrow \mathcal{O}$ defined by  $\sigma(\textbf{x}_j, F(\textbf{x}_j)) =\arg\max_{o_i\in\mathcal{O}} \{\mathcal{C}\} $ and used to prioritise the operators and select the most suitable one. The production of $c_{i,t}$ and the implementation of $\sigma(.)$ appear to be the main areas of focus. That is, these two points are mostly the focus of the research in this area. In order to address the intended problem, the approach proposed in this article uses reinforcement learning techniques.


\subsection{Related Work}
\label{sec:related-work}
This section reviews the state-of-the-art research on adaptive operator selection and reinforcement learning. Adaptive operator selection (AOS) is an emerging paradigm that tries to automatically choose the best reproduction operator in accordance with the most recent search dynamics. 
Since the early 1990s, researchers have investigated adaptive operator selection, focusing on how it applies to evolutionary algorithms~\cite{davis1989adapting}. The past decade has seen a considerable advancement in the development of operator selection schemes, which are employed to enhance the efficiency of evolutionary algorithms used to address multi-objective optimisation problems~\cite{lin2016adaptive,verheul2020influence}. 

The Differential Evolution with Landscape-based Adaptive Operator Selection Algorithm, which chooses optimal differential evolution mutation strategies from a pool of five operators using both a landscape measure and a performance history metric, was introduced in~\cite{sallam2017landscape}. The authors looked into the consequences of the suggested algorithm's parameters, including population size, minimum population size, and cycle length. They have evaluated the proposed algorithm's performance by solving 45 unrestricted optimisation problems. However, without using any structured learning method, the proposed approach selects operators based on their performance and fitness landscape.

In order to better identify the best appropriate crossover operator, the work by~\cite{consoli2016dynamic} introduced an adaptive operator selection mechanism that combines a set of four fitness landscape analysis methodologies with an online learning algorithm called dynamic weighted majority. That is their approach suggests training a dynamic regression model for online learning to develop an operator selection scheme using fitness landscape analysis metrics.

In~\cite{kizilay2020differential}, the authors present a differential evolution method with Q-Learning (DE-QL) for engineering design problems (EDPs). The proposed DE-QL generates the trial population using the QL approach, where the QL guides the choice of the mutation strategy among four different strategies as well as crossover and mutation rates from the Q table. If the (finite) number of states is kept low enough to tabulate each against each action, their approach provides good results. However, working with Q-Table-based learning techniques is not feasible in comparatively larger state spaces.

In a more recent work~\cite{durgut2021adaptive_1}, the authors propose an adaptive operator selection strategy based on reinforcement learning, where the operators are mapped to the problem states according to the operation success rate. The Hard-C-Means clustering method and Q-Learning have been combined to create a reinforcement learning algorithm that helps the search agent learn how to choose the appropriate operator online based on the given situation. The authors have evaluated the proposed method in comparison to existing adaptive selection methods for addressing set union knapsack problems that are formalised in a binary representation. The results show that, in compared to well-known adaptive techniques, the proposed approach provides significantly better performance.

In~\cite{Tian22}, the authors propose another operator selection method based on reinforcement learning for evolutionary multi-objective optimisation problem. Their proposed method suggests potential operators to evolve the population for various multi-objective optimisation problems by utilising deep neural networks to learn the relationships between decision variables and Q values of candidate operators during the optimisation process. The action-value function, which is always defined based on the discount sum of future rewards, may be approximated using the conventional Q-learning method because the action space is discrete. In this case, the reward is based on how much the performance of the offspring improved as a result of the use of a genetic operator. The expected cumulative reward of an action given a state is indicated by each grid in a Q-table that stores the reward associated to these actions. However,  the proposed approach simply considers the decision variables of a single solution as a state. In a different approach, according to Bernoulli Thompson sampling, a Beta distribution is established for each operator in~\cite{Sun20} and updated, after which the operators are chosen based on the rewards sampled from the distributions. In this approach , the dynamic Thompson sampling is used as the foundation for adaptive operator selection.

\section{Reinforcement Learning in Operator Selection}\label{sec:RL-OS}
The impact that \emph{operator selection problem} has on the effectiveness and success of problem-solving algorithms has drawn researchers' attention for a long time. A bespoke operator selection scheme (OSS) plays the key role in balancing the exploration and exploitation activities throughout the search process. Although a number of operator selection approaches~\cite{thierens2005adaptive,durgut2021adaptive} have been thoroughly investigated, there are still many challenges and open problems to be addressed. Recently, bespoke OSSs have been developed for higher efficiency using RL techniques~\cite{koulouriotis2008reinforcement,durgut2021adaptive_1}. 

Reinforcement learning is an online learning paradigm that implements active learning in a more structured way, especially when data is either not available or is only temporarily available. Obviously, it is challenging to gather search data and rely on it to create systems for data-driven decision-making. Operator selection is a procedure that is integrated into search operations and makes a selection decision based on inputs from the search environment. This enforces the adoption of reinforcement learning in order to empower operator selection using the data-driven techniques.


One of the well-known RL algorithms is Q learning, which takes $s\in\mathcal{S}$ as the current state of a problem and estimates $Q(s,a)$, the level of the quality measure, once action $a\in \mathcal{A}$ is applied to the problem state, where $Q:\mathcal{S}\times\mathcal{A} \rightarrow\mathbb{R}$. Once the action $a$ is performed in the state $s$, a new state $s'$ is generated, and the expected quality of the new state $s'$ is estimated using $E(Q(s',a))$ as a stochastic function subject to the selected policy, which is often implemented as: 
 
\begin{equation}
\label{eq:estimate}
    E(Q(s',a))=\arg\max_{a\in\mathcal{A}}Q(s',a)
\end{equation}    

Following each action $a$ taken over state $s$, a Bellman equation-based implementation value iteration is used to update the corresponding quality level, and it is represented as:

\begin{equation}\label{eq:update}
    Q(s,a) = Q(s,a)+\beta\Bigl(r+\gamma Q(s',a)-Q(s,a)\Bigr)
\end{equation}


\noindent where $r$ is the reward generated to evaluate/reward the action $a$, $\beta$ is the learning rate, and $\gamma$ is known as the  discount factor. It is important to note that $r$ is the credit --positive or negative-- estimated using a reward function, $R(s,a)$, and the objective of an RL algorithm is to choose actions that maximise the agent's expected cumulative reward.

The estimation of $Q(s,a)$ is handled via a data structure called the Q-Table, where $Q$ values are tabulated in cells of a table of $\mathcal{S}\times\mathcal{A}$. Whenever the cell of $(s,a)$ is activated, the relevant $Q$ value is updated using the Equation~\ref{eq:update}~\cite{watkins_dayan_1992}. However, this approach is not scalable or sufficiently dynamic. There have been other proposed approximation methods, such as feed-forward neural networks~\cite{tesauro_1995} and clustering algorithms~\cite{aydin2000dynamic}. The Hard-c-Means algorithm was used by the authors of~\cite{aydin2000dynamic} to set up cluster centres in conjunction with reinforcements, and they have developed a quality estimation method based on $\|s-c_a\|$, the distance between state $s$ and $c_a$ as the cluster centre for action $a$, where both $s\in\mathcal{S}$ and $c_a\in\mathcal{C}$ represent multi-dimensional data points of the same size (i.e. $dim(s)=dim(c_a)$), $|\mathcal{C}| = |\mathcal{A}|$ and $c_a$ is the cluster centre allocated to the action $a$. 

The clustering algorithm works as follows. First of all, $|\mathcal{A}|$ number of clusters are established, and every cluster centre is initialised to $0$. It runs alongside Q learning and the agent can choose the action with the best fit $a$, apply it to the current state $s$, and generate $s'$ accordingly. In this setting, if the action $a$ results in higher quality, state $s$ will be categorised as $c_a$, the cluster for action $a$. The centre for cluster $c_a$ will be updated accordingly. The $Q(s,a)$ value will be updated using the Equation~\ref{eq:update}. Given that both $s=\{s_i|i=1\ldots dim(s)\}$ and $c_a=\{c_{a,i}|i=1\ldots dim(c_a)\}$ are two vectors of the same size, the centre is updated as:
\begin{equation}\label{eq:centerino}
    c_{a,i}=\frac{c_{a,i}+s_i}{n_a} \quad   \forall i\in \{1,\dots, dim(s)\}
\end{equation}
where $n_a$ is a counter that counts how many times action $a$ has been chosen correctly. As the algorithm iterates, the $Q(s,a)$ values are retrieved with $Q(s,a)=\|s-c_a\|$ as the distance between the state $s$ and the cluster centre $c_a$, where the more data flows in the cluster centres are further fine-tuned.

The operator selection problem, where the problem is to choose the best operator for the given situation, has been introduced in Section~\ref{sec:os}. Given the set of operators $\mathcal{O}$ to be selected for the current state of the problems, $\textbf{x}_t$ subject to the credit level used by the selection scheme $\sigma(.)$. In order to implement the RL algorithm explained above, the set of operators is adopted as the action list $\mathcal{A}=\mathcal{O}$, and the selection criteria is adopted to be $Q(\textbf{x}_j, o_i)$ retrieved from the corresponding cluster centre $Q(\textbf{x}_j, o_i)=\|\textbf{x}_j - c_{o_i} \|$. Therefore,

\begin{equation}\label{eq:select}
\sigma(\textbf{x}_j, F(\textbf{x}_j)) =
\begin{cases}
     \arg\max_{o_i\in\mathcal{O}} Q(\textbf{x}_j, o_i) & \text{if  }rnd\geq\epsilon \\[6 pt]
     \text{random}(o_i\in\mathcal{O}) & \text{otherwise}
\end{cases}   
\end{equation}
where $rnd \in \mathbb{R}$ is a random value in $[0,1]$, $\text{random}(.)$ is random-selection function and $\epsilon$ is a threshold -- usually $\epsilon\leq 0.2$ -- setup for an $\epsilon$-greedy policy to keep exploration and exploitation harmonised through the learning process. 

One of the key elements of reinforcement learning algorithms is reward generation, which determines how the environment will react to the actions the learning agent takes. In this case, the operator selection scheme is the learning agent since it learns how to choose the best operator under the given circumstances. The main aim is to maximise the accumulated reward, $\mathcal{R}=\sum_{t=1}^\infty r_t$, where $r_t$ is the reward generated at iteration $t$ for one of the operators. In fact, $r_t$ is defined to be $r_t=R(s,a)$, which needs to be implemented for the operator selection problems. Each operator, $o_i\in\mathcal{O}$, is required to accumulate and maximise its reward, $r_{o_i,t}$, which transforms $r_t$ into a vector of rewards, $r_t=\{r_{o_i,t}|i=1,\dots,|\mathcal{O}|\}$, and $r_{o_i,t}=R(\textbf{x}_j, o_i)$. 

The reward per action is measured as a normalised difference between the quality of the solutions of the state undertaken, $\textbf{x}_j$, and the state generated with the selected operator, $\textbf{x}^{'}_j=o_i(\textbf{x}_j)$, where it is defined as:


\begin{equation}
    F(\textbf{x}_j)=\frac{\Delta f\times f^{*}}{f(\textbf{x}^{'}_j)}
\end{equation} 
where $f^{*}$ is the globally known best fitness, $f(\textbf{x}^{'}_j)$ and $f(\textbf{x}_j)$ are the quality of the solutions of $\textbf{x}^{'}_j$ and $\textbf{x}_j$, respectively and $\Delta f = f(\textbf{x}^{'}_j)-f(\textbf{x}_j)$. The normalised difference can be negative or positive with respect to the level of improvement. 


\subsection{The Problem State Representation and Feature Analysis} 
\label{sec:features}
A binary approach has been used to express the problem states, with the input set being represented as a collection of binary variables. This approach appears to be not scalable with respect to the size of the problems. In order to overcome this issue, further research has been conducted to identify the representative information captured from fitness landscape and other relevant population-wise information. 

Fitness landscape analysis has been studied over decades to help identify the properties of the solution space that can lead to a highly effective search. In particular, the key details of the locality of the solutions can be derived from the fitness landscape analysis. Over the past few decades, a significantly sound literature has been developed for utilising the most representative information~ \cite{fragata2019evolution, ochoa2019recent, pitzer2012comprehensive}.

The literature has yielded a number of features that can be used to express the problem states in a more scalable manner. This supports in avoiding earlier issues and helps make use of the experiences gained when solving a range of problems of different sizes and types. Let $\Phi=\{\phi_i|i=1\ldots|\Phi|\}$ represent the set of features that can be used in problem characterisation, where $|\Phi|$ is the size of the feature set, which in this study is considered to be $19$. 

These features are partially taken from the properties of the population of solutions and partially extracted from the properties of the individual solutions.  The list of metrics, which are to be considered as features, extracted from the population properties is tabulated in Table~\ref{tab:feature_pop}, whereas the list extracted from individual properties is tabulated in Table~\ref{tab:feature_indvl}. The Appendix contains a list of equations Eq.~\ref{eq:phi_1} -- \ref{eq:phi_19}, which represent the corresponding calculation details.
 

The features listed in Table~\ref{tab:feature_pop} are population-based metrics. In this study, the Artificial Bee Colony Algorithm (ABC) was chosen as the swarm intelligence framework. The first $5$ metrics, $\{\phi_{1},..,\phi_{5}\}$, were obtained from~\cite{teng2016self} and implemented for (i.e. tailored to) the ABC. The ABC is one of the very recently developed highly reputed swarm intelligence algorithms~\cite{karaboga2014comprehensive}. 
Hamming distance has been used to binarise the metrics that were calculated using the distance measure. This has been investigated by~\cite{erwin2020diversity} in order to adjust them to binary problem solving approach. The metrics, $\{\phi_{6},..,\phi_{9}\}$ are introduced and proposed by~\cite{wang2017population} with sound demonstration, while $\phi_{10}$ is obtained from the trail index used in ABC and utilised to measure/observe the iteration-wise hardness in problem solving. In addition, $\phi_{11}$ has been used from~\cite{anescu2017fast} to calculate the distance between the two farthest individuals within a population/swarm. According to~\cite{erwin2020diversity}, one of the crucial factors in swarms/populations that can help characterise the states is the diversity of populations, whereas~\cite{wang2017population} discusses the evolvability of populations in environments with dynamic landscape structure.

On the other hand, a number of metrics, or features, can be obtained from the auxiliary information of each individual solution. These characteristics appear to be useful in elements that are specific to each individual and allow the operators to take meaningful action in specific situations.

The individual-related features are tabulated in Table~\ref{tab:feature_indvl}. The majority of these features were proposed by~\cite{teng2016self} with the exception of $\phi_{18}$, which is introduced for the first time in this study. Among these features, the success rate for operator $i$ is calculated with ${osr}_i = \frac{{sc}_i }{{tc}_i} $, labelled as $\phi_{19}$, where $sr_i$ is the success counter and $tc_i$ is the total usage counter for the operator $i$. 

\begin{table}[htb!]
\caption{ Population-based features}\label{tab:feature_pop}
\centering
\scriptsize
\begin{tabular}{l|l}
\hline
\textbf{Feature}                      & \textbf{Formula} \\ \hline
Population Solution Diversity ($\phi_1$)           & Eq:~\ref{eq:phi_1} \\ 
Population Fitness Deviation ($\phi_2$)            & Eq:~\ref{eq:phi_2} \\ 
Population of new best children ($\phi_3$)         & Eq:~\ref{eq:phi_3} \\ 
Proportion of new improving children ($\phi_4$)    & Eq:~\ref{eq:phi_4} \\ 
Proportion of amount of improvements ($\phi_5$)    & Eq:~\ref{eq:phi_5} \\ 
Proportion of convergence velocity ($\phi_6$)      & Eq:~\ref{eq:phi_6} \\ 
Proportion of convergence reliability ($\phi_7$)   & Eq:~\ref{eq:phi_7} \\ 
Evolutionary ability of population ($\phi_8$)      & Eq:~\ref{eq:phi_8} \\ 
Evolvability of population ($\phi_9$)              & Eq:~\ref{eq:phi_9} \\ 
Proportion of average trial number ($\phi_{10}$)   & Eq:~\ref{eq:phi_10} \\ 
The diameter of population   ($\phi_{11}$)         & Eq:~\ref{eq:phi_11} \\ \hline
\end{tabular}
\end{table}

\begin{table}[!ht]
\scriptsize
\caption{Individual solution-based features}\label{tab:feature_indvl}
\centering
\begin{tabular}{l|l}
\hline
\textbf{Feature}                      & \textbf{Formula} \\ \hline
Distance between $g_{best}$ and parent solutions ($\phi_{12}$)   & Eq:~\ref{eq:phi_12} \\ 
Distance between parent and child solutions ($\phi_{13}$)        & Eq:~\ref{eq:phi_13} \\ 
Fitness gap between $g_{best}$ and child solutions ($\phi_{14}$) & Eq:~\ref{eq:phi_14} \\ 
Fitness gap between the parent and the offspring ($\phi_{15}$)   & Eq:~\ref{eq:phi_15} \\ 
Distance between $p_{best}$ and parent solutions ($\phi_{16}$)   & Eq:~\ref{eq:phi_16} \\ 
Distance between $p_{worst}$ and parent solutions ($\phi_{}17$)  & Eq:~\ref{eq:phi_17} \\ 
Proportion of trial number  ($\phi_{18}$)                        & Eq:~\ref{eq:phi_18} \\ 
The success of operator $i$ ($\phi_{19}$)                        & Eq:~\ref{eq:phi_19} \\ \hline

\end{tabular}
\end{table}

There are more metrics reported in the literature that are calculated through local search procedures \cite{fragata2019evolution, macias2019application}. The set of features listed or tabulated in this article is restricted due to the scope of the study, leaving out many others. To characterise the search space and describe problem states in order to achieve the mapping of problem states to operators, these elicited metrics have only previously been employed separately or in different groups. Several supervised machine learning algorithms have been used in previous work to investigate the impact of this set of features~\cite{durgut2022analysing}. That investigation gave out a lot of positive results regarding the utilisation of the same feature set for capturing instant changes in online search and identifying relevancies to build up experience and apply it for instant decision-making. Given that a generic problem state is denoted by $\textbf{x}_t$ at iteration $t$, each problem state subject to a decision will be generated by $\Phi_t = g(\textbf{x}_t)$, where $\Phi_t=\{\phi_{t,i}|i=1,\dots,|\Phi_t|, t=1,\dots, T\}$, $g(\textbf{x}_t)$ is a function of $\textbf{x}_t$ that runs all Eq:~\ref{eq:phi_1} - \ref{eq:phi_19} to obtain $\Phi_t$ vector, and finally, $Q(\Phi_t, o_a)=\|\Phi_t - c_{a} \|$. Therefore,


\begin{equation}\label{eq:select_1}
 \sigma(\Phi_t , F(\textbf{x}_t))=
\begin{cases}
     \arg\max_{o_a\in\mathcal{O}} Q(\Phi_t, o_a) & \text{if  }rnd\geq\epsilon \\[6 pt]
     \text{random}(o_i\in\mathcal{O}) & \text{otherwise}
\end{cases}   
\end{equation}
Eq:~\ref{eq:centerino} will be implemented with Eq:~\ref{eq:centerino_1} in order to update the cluster centre corresponding to operator $o_a\in \mathcal{O}$ as the operator selected. 

\begin{equation}\label{eq:centerino_1}
    c_{a,i}=\frac{c_{a,i}+\phi_{t,i}}{n_a} \quad   \forall i\in \{1,\dots,|\Phi_{t}|\}
\end{equation} 
where $c_{a,i}$ is the $i^{th}$ dimension of the cluster centre representing operator $o_a$, $a$ is the operator index, $n_a$ is the number of successful use of the operator $o_a$, $t$ is the iteration index and $i$ is the dimension index for the state data of $\Phi_{t}$.

\subsection{The Search Space Characterisation}
\label{sec:search_space}
It is well-know that the level of difficulty eventually tightens through progressing search, and problem solving is a set of actions taken to find better and best fitting solutions. This indicates that moving to a better fitting solution is very difficult in the later stages but highly likely in the earlier stages. Obviously, this has an impact on how well operators perform, as any operator can contribute to fitness improvements in the early stages but only the fittest operators can help improve the quality of the solution. Due to this fact, the search space $\mathbb{S}$ is decided to be splitted into a number of sub-spaces, $\mathrm{S}_i\in\mathbb{S}$, where $\mathbb{S}=\{\mathrm{S}_i|i=1,..,M\}$ and $M$ is the number of sub-spaces. 

The characteristics of each search space will impose different behaviours for each operator, hence, it is expected that the performance of the operators would be affected accordingly. Because of this, it is necessary to redefine how operators are used so that the experiences obtained in each sub-space can be taken into account, recorded independently, and utilised effectively. The set of operators defined as $\mathcal{O}=\{o_a|a=1,..,|\mathcal{O}|\}$ will be redefined as $\mathcal{O}=\{o_{a,i}|a=1,\dots,|\mathcal{O}|, i=1,\dots, M\}$. In order to avoid confusions and congestion of indexes, a proxy set of operators, $\mathrm{O}$, is defined as $\mathrm{O}=\{\mathrm{o}_k|k=1,\dots, K \}$, where $K=|\mathcal{O}|\times M$. Subsequently, $Q$ values will be $Q(\Phi_t, \mathrm{o}_k)=\|\Phi_t - \mathrm{c}_k \|$ and the Eq:~\ref{eq:select_1} and \ref{eq:centerino_1} will be revised as follows:
\begin{eqnarray}
   \sigma(\Phi_t , F(\textbf{x}_t)) &=&
   \begin{cases}
     \arg\max_{\mathrm{o}_k\in\mathrm{O}} Q(\Phi_t, \mathrm{o}_k) & \text{if  }rnd\geq\epsilon \\[6 pt]
     \text{random}(\mathrm{o}_k\in\mathrm{O}) & \text{otherwise}  
   \end{cases}\label{eq:select_2}\\ [8pt]
    \mathrm{c}_{k,i}&=&\frac{\mathrm{c}_{k,i}+\phi_{t,i}}{\mathrm{n}_k} \quad \quad  \forall i\in \{1,\dots,|\Phi_{t}|\} \label{eq:ceterino_2}
\end{eqnarray}
where $\mathrm{c}_{k,i}$ is the $i^{th}$ dimension of the cluster centre representing operator $k$, $\mathrm{n}_k$ is the number of successful use of the operator $k$, $t$ is the iteration index and $i$ is the dimension index for the state data of $\Phi_{t}$. 

\subsection{Transfer Learning using Reinforcement Learning}
\label{sec:transfer}
Transfer learning is a sub-field of machine learning where the learned experience is formulated for use in processing unseen data and information and solving new problems. Data modelling is the process of instrumenting algorithms to device machinery that processes unseen data and projects prospective cases. The goal of transfer learning is to increase the effectiveness of the algorithms by processing and translating the experience obtained during previous runs and episodes while solving one or more problems into a transferable form. In this study, transfer learning has been used to apply experience gained through RL to a range of problem instances, both large and small and of various types. 

Transfer learning via RL is a relatively unexplored area of research.  This approach looks at how knowledge and experience gained through online learning can be transferred into a format that can be used to solve similar problems. Although it is well known that a trained data model can be applied to similar cases with unseen data and the performance can be significantly improved~\cite{Lisa2010}, it is not clear how this can be utilised when the problem or domain is entirely different and the interrelation of the components are also different.
Traditional machine learning and deep learning algorithms have up until now been designed to work in isolation \cite{fragata2019evolution, macias2019application}. These algorithms have been designed to solve specific problems. Once the feature-space distribution changes, the models must be rebuilt from scratch.

Transfer learning turned to be very popular in machine learning that are proposed to overcome the isolated learning paradigm. It enables use of gained knowledge and experience over the previous times while solving other problems previously. The idea is to transfer the gained experience from solving one problem to the other related problems. The transfer learning process is required to address the following three critical questions: what should be transferred, when should it be transferred, and how should it be transferred? The literature includes a number of transfer learning approaches that can be used, depending on the domain, the problem at hand, and the data availability~\cite{Sinno2010}. It is essential to identify the similarities and differences between the problems, which happened to be the problem with which the knowledge to be used gained through in order to solve the problem at the hand. The characteristics of such problems play crucial role in success of the knowledge transferred to be helpful in problem solving undergoing. This turns to gain more importance when RL agents are used in transfer learning. The majority of transfer learning research has been focused on general classical RL problems; however, the proposed work will focus on how to gain transferable experience in \emph{operator selection} through reinforcement learning.

The experience gained can be transferred via a data model defined as $\mathcal{M}:\mathbb{R}^{|D|}\rightarrow \mathbb{R}^N$, where $D$ is the data set in hand, $d\in D$ is a piece of data, $N$ is the number of outputs of the model, and the instantiated model is defined as an accumulated structure trained through: 
\begin{equation}
\label{eq:union}
  \mathcal{M}(D)=\bigcup_{\substack{i\in |\mathcal{M}|\\d\in D}}^{\infty}\lambda_i m_i(d)  
\end{equation}

It is known that the instantiated data model, $\mathcal{M}(D)$, is a union of training episodes, $m_i(d)$, which is an episodic implementation of $\mathcal{M}(D)$, and is discounted/regulated by the coefficient $\lambda_i$. Each upcoming training episode $m_i{d}$ is aggregated to the model $m(d)$ normalised with $\lambda_i$. The aggregations are anticipated to increase the learning agent's degree of experience, subject to a decision over whether to grant the agent permission to accumulate experience in a single run or in an episode.

The Eq:~\ref{eq:union} rule represents the longer run of transfer learning logic. However, this model require to be implemented or instantiated in order to be simplified and implemented as $m(d)=(1-\delta){\alpha}(d) + \delta {\beta}(d)$, where $\alpha(d)$ and $\beta(d)$ represent two successive training episodes, $m_i(d)$ and $m_{i+1}(d)$, respectively. While the learning coefficients $\lambda_{i+1}=\delta$ and $\lambda_i=(1-\delta)$ control the contribution of previous and next experiences, respectively. For example, it switches training ON if $\delta>0$ and OFF otherwise. 

The above-mentioned method for incorporating transfer learning into problem solving employs the data model as an ongoing clustering approach, where the model $ \mathcal{M}(D)\xleftarrow[]{} \mathcal{C}(D)$ is instantiated and trained with solving a specific instance of the problem that includes the following components: $m_i(d)\xleftarrow[]{}\textbf{c}_a(d)$, where $\alpha$ represents the learned components, $\beta$ is the change that will be brought about by impending actions, and $\delta$ is the determination of whether or not past experience will be used.

The ABC algorithm is implemented with a learning operator selection scheme, and it is designed to run once to solve a particular problem instance. Online learning is turned ON to train and optimise the cluster centres that represent the operators, and it is turned OFF to repeat the experiments using the same problem instance but different random number sequences. Due to the pruning of the exploration activities, it is anticipated that the problem will be solved in a much shorter amount of time with similar or slightly improved solution quality. This stage, which is still the most popular way of transfer learning, is where the proposed algorithm transfers previously gained experience.

The transfer learning implemented in this study is aimed to be employed in three layers of transfer experience: (i) The agent is trained once by solving a specific benchmark instance of one of the combinatorial problems (such as OneMax or the Set Union Knapsack Problem), and then using that experience to solve several other instances of the same size; (ii) using the trained agent in solving various other benchmark instances of the same problem type that are of different size from the original instance; and (iii) using the trained agent to solve the instance of the other combinatorial problem being studied in this paper. The preliminary results of the layer (i) of this approach with limited depth have been reported in~\cite{durgut2022transfer}. In this article, we provide further details of the results with relevant discussions in the following section.

\section{Experimental Results}\label{sec:expr-results}
%
This section presents the experimental results and evaluates the performance of the proposed algorithm which is coded using Python 3.10 programming language. All the experiments were conducted on the compute server provided by TUBITAK ULAKBIM, High Performance and Grid Computing Center. Two well-known combinatorial optimisation problems, which are discussed in the following subsection, have been solved using the proposed approaches. 
 
\subsection{Benchmark Problems}
The algorithms proposed are implemented to demonstrate how the proposed approaches work and offer better solutions. For that purpose, two renowned combinatorial optimisation problems have been selected and a number of benchmarking instances have been generated and used for the demonstration accordingly. The selected two problems are \textit{OneMax} and \textit{Set Union Knapsack problems}, which are widely acknowledged and well known in various respects.
   
\textit{OneMax }problem is a simple binary problem, which mainly looks for the highest value that can be gained from a set of strings. Let $\mathcal{B}$ be the set of binary values and $b_i\in \mathcal{B}$ is a particular binary value. The main aim is to maximise $\sum_{i=1}^{|\mathcal{B}|}b_i$. The problem has been used to demonstrate the performance of a genetic algorithm by \cite{khair_2018}. A number of benchmarking problem instances have been generated with a range of dimensions from $500$ to $5000$ as tabulated in Table \ref{table:OneMax}.

\begin{table}[h]
\centering
\scriptsize
\caption{OneMax problem instances}
\label{table:OneMax}
\begin{subtable}[h]{\textwidth}
\begin{tabular}{l|cccccccccc}
Problem ID & 1   & 2   & 3    & 4    & 5    & 6    & 7    & 8    & 9    & 10  \\ \hline
Dimension  & 500 & 750 & 1000 & 1250 & 1500 & 1750 & 2000 & 2250 & 2500 & 2750 \\
\end{tabular}
\end{subtable}
\vspace{10pt}

\begin{subtable}[h]{\textwidth}
\begin{tabular}{l|cccccccccc}
Problem ID & 11   & 12   & 13   & 14   & 15   & 16   & 17   & 18   & 19 & $\quad$ \\\hline
Dimension  & 3000 & 3250 & 3500 & 3750 & 4000 & 4250 & 4500 & 4750 & 5000 & $\quad$\\
\end{tabular}
\end{subtable}
\end{table}

\textit{Set Union Knapsack problem} (\textit{SUKP}) is another combinatorial optimisation problem, which is usually solved as binary optimisation problem. It is known as one of the NP-Hard problems \cite{goldschmidt1994note}, which is attempted by many metaheuristic approaches \cite{ozsoydan2019swarm, he2018novel, wu2020solving}. It requires a set of items to be optimally composed in subsets so as to gain the maximum benefit. Given a set of $n$ elements, $U=\{u_i|i=1,\ldots,n\}$ with a non-negative weight set, $W=\{w_i|i=1,\ldots,n\}$ and a set of $m$ items, $\mathcal{K}=\{U_j|j=1,\ldots,m\}$ with a profit set, $\mathcal{P}=\{p_j>0|j=1,\ldots,m\}$, a subset of $L\subseteq \mathcal{K}$ is sought to be found such that it maximises the profit subject to the sum of the weights of the selected items not exceeding the capacity constraint, $C$. 

The problem is originally represented in real numbers and needs to be represented in binary form to enable binary operators in search algorithms such as binary ABC \cite{durgut2021adaptive}. Following the details of the problem and the approach introduced by \cite{wu2020solving}, a binary vector, $B=\{b_j|j=1,..,m\}\in \{0,1\}$, is defined to be used as the set of decision variables, where $b_j=1$ if an item is selected, $b_j=0$, otherwise. The model of the problem can be reformulated as follows:

\begin{equation}
    \max \quad f(B) =  \sum_{j=1}^{m}{b_jp_j}
\end{equation}
\begin{equation}
    \text{s.t.} \quad W(L_B) =  \sum_{i\in \bigcup_{j \in L_B} U_j}{w_i \leq C }
\end{equation}

The main goal is to find the best binary vector, $B$, which provides the subset of items with the maximum profit. 

The problem instances of \textit{SUKP} chosen in this study are collected from recently published literature. In \cite{he2018novel}, the authors have introduced $30$ benchmarking problem instances of \textit{SUKP} as tabulated in Table \ref{tab:SUKPinstances} with all configuration details, where there are $3$ different configurations that vary depending on the comparative status of $m$ and $n$: (i) $m>n$, (ii) $m<n$, and (iii) $m=n$, while $\textbf{w} \in \{0.10, 0.15\}$  and $\textbf{y} \in \{0.75, 0.85\}$ represent the density of elements and the rate between the capacities and the sum of weights of elements, respectively. As shown, each set of problem instances includes $10$ instances that vary in $m, n, \textbf{w}$ and $\textbf{y}$ values. 

\begin{table}
\scriptsize
\caption{The \textit{SUKP} benchmark instances introduced in the literature \cite{he2018novel} }
\label{tab:SUKPinstances}
\centering
\smallskip\noindent
\resizebox{\linewidth}{!}{%
\begin{tabular}{rrrrr|rrrrr|rrrrr}
\hline
\multicolumn{5}{c|}{$\text{Set}_1$}     & \multicolumn{5}{c|}{$\text{Set}_2$}   & \multicolumn{5}{c}{$\text{Set}_3$}   \\ \hline
ID        & m & n & w & y & ID        & m & n & \textbf{w} & \textbf{y} & ID        & m & n & \textbf{w} & \textbf{y}  \\ \hline
1\_1  & 100   & 85  & 0.10  & 0.75  & 2\_1  & 100  & 100   & 0.10  & 0.75  & 3\_1  & 85  & 100   & 0.10   & 0.75    \\
1\_2  & 100   & 85  & 0.15  & 0.85  & 2\_2  & 100  & 100   & 0.15  & 0.85  & 3\_2  & 85  & 100   & 0.15   & 0.85    \\
1\_3  & 200   & 185 & 0.10  & 0.75  & 2\_3  & 200  & 200   & 0.10  & 0.75  & 3\_3  & 185  & 200  & 0.10   & 0.75    \\
1\_4  & 200   & 185 & 0.15  & 0.85  & 2\_4  & 200  & 200   & 0.15  & 0.85  & 3\_4  & 185  & 200  & 0.15   & 0.85    \\
1\_5  & 300   & 285 & 0.10  & 0.75  & 2\_5  & 300  & 300   & 0.10  & 0.75  & 3\_5  & 285  & 300  & 0.10   & 0.75    \\
1\_6  & 300  & 285  & 0.15  & 0.85  & 2\_6  & 300  & 300   & 0.15  & 0.85  & 3\_6  & 285  & 300  & 0.15   & 0.85    \\
1\_7  & 400  & 385  & 0.10  & 0.75  & 2\_7  & 400  & 400   & 0.10  & 0.75  & 3\_7  & 385  & 400  & 0.10   & 0.75    \\
1\_8  & 400  & 385  & 0.15  & 0.85  & 2\_8  & 400  & 400   & 0.15  & 0.85  & 3\_8  & 385  & 400  & 0.15   & 0.85    \\
1\_9  & 500  & 485  & 0.10  & 0.75  & 2\_9  & 500  & 500   & 0.10  & 0.75  & 3\_9  & 485  & 500  & 0.10   & 0.75    \\
1\_10 & 500  & 485  & 0.15  & 0.85  & 2\_10 & 500  & 500   & 0.15  & 0.85  & 3\_10 & 485  & 500  & 0.15   & 0.85    \\ \hline
\end{tabular}}
\end{table}

The Artificial Bee Colony (ABC) algorithm considered in this study has been implemented to embed pool of operators with an operator selection scheme. As explained before, the selection scheme is developed and empowered with reinforcement learning so as to automatise the whole mechanism. The ABC framework has three main parameters to setup; maximum iteration, the number of solutions in the population, and the number of maximum trials (limit). For the \textit{OneMax} problem, the swarm/population size is set to $N = 20$ and the maximum iteration is fixed at $250$, whereas for \textit{SUKP} cases, the swarm/population size is determined by the dimensionality of the problem instances, with $max\{m, n\}$, where $m$ is the number of items and $n$ is the number of elements in an instance.

The pool of operators consists of four binary operators, which have recently been used in the literature, $\mathcal{O}=\{flip\text{ABC},n\text{ABC}, ibin\text{ABC}, nb\text{ABC}\}$.  \textit{flip}ABC simply inverts a randomly selected bit, while the others -- \textit{n}ABC, \textit{ibin}ABC and \textit{nb}ABC -- are more complex-structured operators taken from \cite{xiang2021artificial}, \cite{durgut2021improved} and \cite{santana2019novel}, respectively. According to the conditions of the search algorithm, an operator selection scheme is aimed to function for selecting the most productive operator from this pool.   

Four variants of the Reinforcement Learning (RL)-based adaptive operator selection schemes embedded in ABC have been tested. The first variant is labelled with \textit{One-Run} that disregards previous learning experiences and starts learning from scratch in each repetition re-initialising all parameters of the algorithm.  On the other hand, the variant labelled with \textit{All-Run} does not dispose gained experience up-to-date and transfers all relevant operator data to the next runs. The variant labelled as \textit{One-Run w/L} has uploaded information previously recorded on the specific problem instance (for the \textit{OneMax} problem $2500$ dimensions, for the \textit{SUKP} ID X) after $30$ runs. The variant labelled as \textit{All-Run w/L} has the same idea but only upload the information for the first run, because the information is transferred to the next run. 

\subsection{Results and Discussions }
The experimental results presented in the following sections was conducted to evaluate the efficiency of problem representation using the set of features introduced in Section \ref{sec:features}, splitting search space approach as in Section \ref{sec:search_space}, and using the transfer learning detailed in Section \ref{sec:transfer}. The efficiency gained by splitting the search space was first investigated since the rest of the experiment is based on those results.


\subsubsection{Splitting search space}

The search space of a problem can hold different characteristics in different regions since operators' behaviours can vary region-to-region. Therefore, in order to enable the operator selection scheme to learn how to select the operators according to different problem states and stages of evolution, we decided to divide the search space into a number of sub-spaces.
\begin{figure}[ht]
    \centering
    \includegraphics[width = 0.6\textwidth]{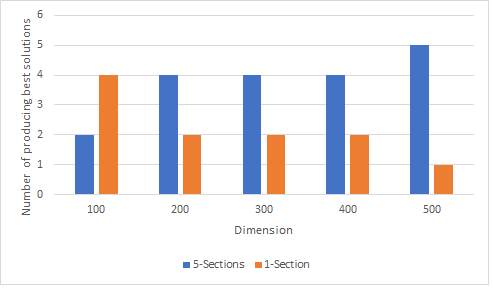}
    \caption{Experimental results for splitting search space into pieces and treating each as an  independent space}
    \label{fig:splitts} 
\end{figure}

The new pool of operators turns to be $\mathrm{O}=\{flip\text{ABC}_i,n\text{ABC}_i, ibin\text{ABC}_i, nb\text{ABC}_i|i=1,\dots,5\}$, as an extension of $\mathcal{O}$. Here, the same pool of operators has been applied to each sub-space separately without reusing any  experience data gained from previous (other) sub-spaces. Therefore, the adaptive operator selection scheme is trained to learn when to select each operator subject to the circumstances of each sub-space. 

Figure~\ref{fig:splitts} present comparative results in bar chart considering two options; the complete search space labelled with \textit{1-section}, and a split of 5 sub-spaces labelled with \textit{5-section}. On the x-axis, the dimensions of \textit{OneMax} problem instances plotted against the quality of the solution on y-axis. Each x-point represents 6 different instances in the same dimension, where the quality of the solution is attained much better in higher dimension with 5-section splits noting that 1-section approach achieves higher than 5-section in 100-dimension problems only. This clearly demonstrates the fact that splitting search space helps to achieve better performance.    

\subsubsection{Feature-based problem representation}
The problem has been represented with a set of $19$ features which is introduced earlier. The idea is to demonstrate that a good quality of learning can be achieved with a set of latent features – calculated accordingly – in order to make our proposed approach more scalable compared to the binary representation suggested by  \cite{durgut2021adaptive_1}. The approach of \cite{durgut2021adaptive_1} is strictly specific to the problem size, and the results are encouraging but not scalable. In order to overcome these limitations, in our proposed work in this paper, feature-based representation is adopted and implemented so that the size of the problem does not impose any restrictions. 

Table \ref{tab:OneMax_1} and \ref{tab:sukp_1} exhibit the performance of the proposed approaches when applied to solve the \textit{OneMax} and \textit{SUKP} benchmark problems. The tables illustrate the performance metrics with columns labelled "Rank," "Max," "Mean," and "Std" to denote the rank of the algorithm in terms of mean values, its highest-ever fitness value, the mean, and the standard deviation of the repeated tests, respectively. The measures are calculated over $30$ repetitions. The column labelled with 'p-value' denotes the p-value obtained by Wilcoxon signed-rank test. 


Table \ref{tab:OneMax_1} depicts the comparative results of selections by random operator (\textit{Random Selection}) embedded in the algorithm, \textit{One-Run} and \textit{All-Run}, where \textit{All-Run} has obtained the first rank for $11$ instances, while \textit{Random Selection} and \textit{One-Run} achieved the first rank only $5$ and $3$ times, respectively. \textit{Random Selection} produced the best solution only in few instances ($500$, $750$, $1000$ dimensions). Overall, \textit{All-Run} attains mean rank of $1.52$ wins the competition as the best performing algorithm. On the other hand, \textit{Random Selection} and \textit{One-Run} have only achieved the mean rank of $2.68$ and $1.78$, respectively. 



\begin{table}[h]
\centering
\caption{Comparative results for \textit{OneMax} problems with three variants}
\label{tab:OneMax_1}
\smallskip\noindent
\resizebox{\linewidth}{!}{%
\begin{tabular}{ccrrr|crrrr|crrrr}
\multicolumn{5}{c}{\textbf{\textit{Random Selection}}} & \multicolumn{5}{c}{\textbf{\textit{One-Run}}} & \multicolumn{5}{c}{\textbf{\textit{All-Run}}}  \\  \hline

\multicolumn{1}{l}{ID} & \multicolumn{1}{c}{Rank} & \multicolumn{1}{c}{Max} & \multicolumn{1}{c}{Mean} & \multicolumn{1}{c|}{Std} & \multicolumn{1}{c}{Rank} & \multicolumn{1}{c}{Max} & \multicolumn{1}{c}{Mean} & \multicolumn{1}{c}{Std} & \multicolumn{1}{c|}{p} & \multicolumn{1}{c}{Rank} & \multicolumn{1}{c}{Max} & \multicolumn{1}{c}{Mean} & \multicolumn{1}{c}{Std} & \multicolumn{1}{c}{p}   \\ 

\hline
1 & 1          & 500  & 499.87  & 0.43            & 3          & 500  & 499.50  & 0.81  & 2.0E-02 & 2          & 500  & 499.80  & 0.40  & 4.8E-01  \\
2 & 1          & 750  & 748.47  & 1.20            & 2          & 749  & 744.77  & 2.29  & 6.4E-06 & 3          & 748  & 744.67  & 2.10  & 3.3E-06  \\
3 & 1          & 993  & 987.70  & 2.61            & 2          & 989  & 984.03  & 3.38  & 2.9E-04 & 3          & 990  & 983.60  & 3.48  & 4.9E-05  \\
4 & 3          & 1221 & 1209.83 & 5.48            & 2          & 1224 & 1214.87 & 5.73  & 5.3E-03 & 1          & 1224 & 1216.57 & 4.99  & 1.8E-04  \\
5 & 3          & 1430 & 1413.93 & 6.31            & 2          & 1448 & 1437.20 & 7.18  & 2.5E-06 & 1          & 1454 & 1438.60 & 8.65  & 1.9E-06  \\
6 & 3          & 1629 & 1612.57 & 11.02           & 2          & 1664 & 1646.37 & 8.62  & 1.7E-06 & 1          & 1668 & 1646.60 & 9.61  & 1.7E-06  \\
7 & 3          & 1827 & 1801.33 & 9.86            & 1          & 1861 & 1841.97 & 13.73 & 1.7E-06 & 2          & 1873 & 1841.90 & 13.34 & 1.7E-06  \\
8 & 3          & 2014 & 1987.87 & 13.00           & 2          & 2054 & 2030.77 & 14.57 & 1.7E-06 & 1          & 2058 & 2031.90 & 15.81 & 2.3E-06  \\
9 & 3          & 2195 & 2171.80 & 12.38           & 2          & 2251 & 2211.33 & 19.74 & 3.9E-06 & 1          & 2246 & 2218.33 & 14.97 & 1.7E-06  \\
10 & 3         & 2383 & 2347.70 & 14.16           & 1          & 2435 & 2395.33 & 18.65 & 2.2E-06 & 2          & 2445 & 2395.03 & 19.87 & 5.2E-06  \\
11 & 3         & 2557 & 2524.30 & 16.38           & 1          & 2621 & 2572.20 & 17.42 & 1.9E-06 & 2          & 2608 & 2563.23 & 19.30 & 5.2E-06  \\
12 & 3         & 2723 & 2698.10 & 12.66           & 1          & 2786 & 2747.77 & 18.98 & 1.7E-06 & 2          & 2801 & 2743.90 & 23.20 & 1.9E-06  \\
13 & 3         & 2889 & 2864.23 & 14.79           & 1          & 2937 & 2903.00 & 18.34 & 3.2E-06 & 2          & 2943 & 2902.27 & 20.55 & 1.7E-06  \\
14 & 3         & 3073 & 3034.10 & 22.79           & 2          & 3125 & 3081.50 & 20.47 & 1.7E-06 & 1          & 3157 & 3083.47 & 29.91 & 1.2E-05  \\
15 & 3         & 3245 & 3199.87 & 21.79           & 2          & 3305 & 3250.47 & 24.49 & 1.7E-06 & 1          & 3293 & 3251.60 & 29.76 & 7.0E-06  \\
16 & 3         & 3401 & 3364.97 & 17.68           & 2          & 3462 & 3404.60 & 25.20 & 2.4E-05 & 1          & 3491 & 3418.50 & 31.27 & 4.7E-06  \\
17 & 3         & 3558 & 3524.10 & 15.66           & 2          & 3642 & 3563.90 & 34.78 & 8.9E-05 & 1          & 3621 & 3577.47 & 24.08 & 2.1E-06  \\
18 & 3         & 3723 & 3690.00 & 16.59           & 2          & 3797 & 3739.50 & 28.98 & 7.3E-06 & 1          & 3813 & 3750.00 & 33.75 & 3.5E-06  \\
19 & 3         & 3901 & 3854.57 & 23.29           & 2          & 3951 & 3904.90 & 34.57 & 1.0E-05 & 1          & 3965 & 3911.13 & 25.10 & 2.6E-06  \\ 
\hline
Mean &2.68 &      &         &                 &1.78 &      &         &       &         & 1.52 &      &         &  \\ \hline             
\end{tabular}}
\end{table}

The experimentation with feature-based problem representation has been extended by solving \textit{SUKP} problems. Table \ref{tab:sukp_1} presents the comparative results of random operator selection, \textit{One-Run} and \textit{All-Run} on \textit{SUKP} benchmark instances. It can be observed that \textit{One-Run} has obtained the best results for $17$ out of the $30$ instances, while \textit{Random Selection} and \textit{All-Run} have achieved the first rank in $11$ and $16$ times out of the $30$ instances, respectively. The mean rank achieved by \textit{One-Run} is the best with $1.4$ and it achieved the best mean measure for $21$ out of the $30$ instances. On the other hand, \textit{All-Run} has achieved $2$ for the mean rank, while \textit{Random Selection} hits the third place only with $2.6$ for the mean rank. The difficulty with \textit{SUKP} comes from its multi-modality, where each time different routes to best found results requires much longer runs for better learning to avoid the variations of the routes. 
The results tabulated in this subsection suggest that the features used to represent problem states help in solving the instances significantly better than \textit{Random Selection}, which ultimately helps driving the base-line for the expected performance. Both variants of the RL-based selection schemes have achieved quality level that is comparable with previous works such as \cite{durgut2021reinforcement} and \cite{durgut2022transfer} in terms of the quality of the solutions. However, due to the differences in the set of operators utilised, configuration, and settings, the results cannot be directly compared.

\begin{table}[htb]
\centering
\caption{Comparative results with two selection algorithms against selection for \textit{SUKP} problems}\label{tab:sukp_1}
\smallskip\noindent
\resizebox{1.0\linewidth}{!}{%
\begin{tabular}{lcrrr|crrrr|crrrr}
~ & \multicolumn{4}{c}{\textbf{\textit{Random Selection}}} & \multicolumn{5}{c}{\textit{\textbf{One-Run}}} & \multicolumn{5}{c}{\textit{\textbf{All-Run}}}            \\ 
\cline{1-15}
\multicolumn{1}{l}{ID} & \multicolumn{1}{c}{Rank} & \multicolumn{1}{c}{Max} & \multicolumn{1}{c}{Mean} & \multicolumn{1}{c|}{Std} & \multicolumn{1}{c}{Rank} & \multicolumn{1}{c}{Max} & \multicolumn{1}{c}{Mean} & \multicolumn{1}{c}{Std} & \multicolumn{1}{c|}{p} & \multicolumn{1}{c}{Rank} & \multicolumn{1}{c}{Max} & \multicolumn{1}{c}{Mean} & \multicolumn{1}{c}{Std} & \multicolumn{1}{c}{p}   \\ 
 
\cline{1-15}
1\_1  & 2    & 13044 & 13042.63 & Tem.36     & 1    & 13167 & 13052.20 & 30.68  & 1.0E-01 & 3    & 13044 & 13034.87 & 44.27  & 5.9E-01  \\
1\_2  & 3    & 12130 & 11945.79 & 174.61     & 1    & 12274 & 12076.95 & 65.72  & 1.3E-01 & 2    & 12130 & 11993.63 & 168.86 & 7.0E-01  \\
1\_3  & 3    & 13302 & 13104.13 & 95.74      & 1    & 13340 & 13165.27 & 106.95 & 3.1E-02 & 2    & 13319 & 13126.80 & 109.97 & 4.6E-01  \\
1\_4  & 3    & 13671 & 13230.33 & 144.66     & 1    & 13671 & 13376.13 & 191.36 & 1.2E-03 & 2    & 13671 & 13297.73 & 205.28 & 2.0E-01  \\
1\_5  & 3    & 10768 & 10526.47 & 118.52     & 1    & 10980 & 10588.03 & 160.02 & 1.3E-01 & 2    & 10896 & 10548.63 & 132.44 & 7.0E-01  \\
1\_6  & 1    & 12040 & 11407.47 & 359.16     & 3    & 12012 & 11355.87 & 308.24 & 5.5E-01 & 2    & 12035 & 11383.47 & 278.75 & 5.7E-01  \\
1\_7  & 3    & 10852 & 10512.17 & 179.25     & 1    & 10994 & 10616.67 & 180.01 & 8.8E-02 & 2    & 11244 & 10580.90 & 231.11 & 2.3E-01  \\
1\_8  & 3    & 10168 & 9997.70  & 188.32     & 1    & 10168 & 10055.10 & 137.99 & 3.6E-01 & 2    & 10182 & 10036.23 & 139.19 & 4.9E-01  \\
1\_9  & 2    & 11326 & 11076.90 & 147.47     & 1    & 11426 & 11118.60 & 150.97 & 2.9E-01 & 3    & 11296 & 11042.73 & 127.75 & 4.7E-01  \\
1\_10 & 2    & 9523  & 9158.10  & 174.65     & 1    & 9409  & 9166.57  & 127.46 & 9.4E-01 & 3    & 9486  & 9140.33  & 159.16 & 7.5E-01  \\
2\_1  & 3    & 13963 & 13822.40 & 074.49     & 1    & 14044 & 13850.57 & 79.48  & 2.2E-01 & 2    & 13963 & 13843.27 & 68.49  & 4.9E-01  \\
2\_2  & 3    & 13407 & 13205.27 & 206.90     & 2    & 13407 & 13222.67 & 149.27 & 8.5E-01 & 1    & 13407 & 13278.13 & 120.32 & 2.3E-01  \\
2\_3  & 2    & 12257 & 11716.60 & 255.53     & 1    & 12350 & 11792.00 & 253.90 & 1.7E-01 & 3    & 12242 & 11708.17 & 235.69 & 8.5E-01  \\
2\_4  & 3    & 11800 & 11491.20 & 204.61     & 2    & 11821 & 11550.13 & 257.17 & 2.9E-01 & 1    & 11830 & 11613.43 & 192.09 & 2.0E-02  \\
2\_5  & 3    & 12644 & 12370.40 & 170.61     & 1    & 12644 & 12523.10 & 166.57 & 1.9E-03 & 2    & 12644 & 12417.17 & 228.26 & 4.2E-01  \\
2\_6  & 2    & 10724 & 10592.17 & 177.38     & 1    & 10735 & 10618.93 & 118.84 & 5.9E-01 & 3    & 10731 & 10571.00 & 186.18 & 6.5E-01  \\
2\_7  & 3    & 10998 & 10766.07 & 142.97     & 1    & 11001 & 10832.70 & 112.07 & 1.3E-01 & 2    & 11160 & 10831.37 & 135.25 & 1.4E-01  \\
2\_8  & 2    & 10264 & 9607.23  & 194.20     & 1    & 10089 & 9665.67  & 170.46 & 4.5E-02 & 3    & 10250 & 9600.00  & 208.11 & 6.6E-01  \\
2\_9  & 3    & 10700 & 10568.30 & 106.69     & 1    & 10734 & 10592.90 & 94.68  & 6.1E-01 & 2    & 10784 & 10585.33 & 94.82  & 5.2E-01  \\
2\_10 & 3    & 9926  & 9579.80  & 161.23     & 1    & 10176 & 9651.73  & 219.47 & 9.8E-01 & 2    & 10176 & 9592.83  & 172.32 & 4.5E-01  \\
3\_1  & 2    & 11752 & 11381.63 & 129.65     & 3    & 11579 & 11348.37 & 96.03  & 3.0E-01 & 1    & 12045 & 11409.47 & 201.67 & 8.1E-01  \\
3\_2  & 2    & 12369 & 11865.90 & 240.85     & 3    & 12369 & 11864.07 & 289.56 & 8.5E-01 & 1    & 12369 & 11923.83 & 345.64 & 4.7E-01  \\
3\_3  & 3    & 13463 & 13097.63 & 228.63     & 1    & 13392 & 13141.20 & 174.84 & 3.0E-01 & 2    & 13477 & 13119.73 & 211.00 & 7.5E-01  \\
3\_4  & 3    & 10973 & 10612.30 & 197.76     & 2    & 10920 & 10620.63 & 138.02 & 7.2E-01 & 1    & 10920 & 10623.07 & 118.08 & 8.7E-01  \\
3\_5  & 2    & 11538 & 11123.97 & 144.64     & 1    & 11538 & 11154.93 & 147.54 & 2.5E-01 & 3    & 11538 & 11101.37 & 172.69 & 2.6E-01  \\
3\_6  & 3    & 11041 & 10830.23 & 167.47     & 2    & 11414 & 10941.50 & 218.08 & 2.5E-02 & 1    & 11237 & 10944.23 & 187.32 & 3.4E-02  \\
3\_7  & 3    & 10085 & 9877.20  &  79.89     & 1    & 10068 & 9911.37  & 70.59  & 7.8E-02 & 2    & 10294 & 9898.60  & 100.12 & 4.8E-01  \\
3\_8  & 2    & 9588  & 9285.80  & 154.47     & 1    & 9588  & 9294.77  & 138.33 & 9.8E-01 & 3    & 9532  & 9262.50  & 139.24 & 4.5E-01  \\
3\_9  & 3    & 10720 & 10482.37 &  91.00     & 2    & 10653 & 10494.07 & 84.91  & 5.4E-01 & 1    & 10732 & 10504.37 & 97.57  & 5.8E-01  \\
3\_10 & 3    & 9514  & 9314.20  & 117.77     & 2    & 9514  & 9359.17  & 142.50 & 2.5E-01 & 1    & 9514  & 9380.23  & 136.25 & 4.3E-02  \\ 
\cline{1-15}
 Mean     & 2.6  & ~     & ~        & ~          & 1.4  & ~     & ~        & ~      & ~       & 2.0  & ~     & ~        & ~      &     \\ \hline    
\end{tabular}}
\end{table}

\subsubsection{Utilising transfer learning}
The remaining part of the proposed work involves training the RL agents for new problems and problem instances through transfer learning. The goal is to preserve information gained through solving problems with a given type, size, and setting so that it may be applied to solving problems of a new type, size, complexity, and setting. For this reason, two different variants of RL-agents are created; the variants that use past experience are labelled as \textit{w/L}, and the variants that do not use past experiences but instead begin learning from scratch each time are labelled as \textit{w-L}. These two new additional variants are added to the two main variants, \textit{One-Run} and \textit{All-Run}, as previously mentioned, to provide a total of four variants all together.  

Table \ref{tab:OneMax_2} presents the comparative results of the four variants of RL-based ABC -- as introduced previously -- to solving the \textit{OneMax} benchmarks with respect to 'Rank', 'Max', 'Mean', 'Std', and additionally the 'p-values' from the statistical test. It can be observed that the variants that carrying over the previous experience help in producing better solutions compared to those starting from the scratch. The mean of Wilcoxon ranks of \textit{One-Run w-L} and \textit{One-Run w/L} are $3.21$ and $1.94$, respectively. In addition, the latter variant seems performed better in terms of 'Max' values with score of $14$ out of $20$ instances. Similarly, \textit{All-Run w/L} improves the performance compared to \textit{All-Run w-L} with the mean ranks of $1.94$ and $2.89$, respectively. Apart from the first $6$ ones, the majority of the results are the best for all of the benchmark instances. As the performance of many approaches, including those used in this study, remain quite near to one another, this suggests that these instances are highly competitive.

\begin{figure}
    \centering
    \includegraphics[width=\textwidth]{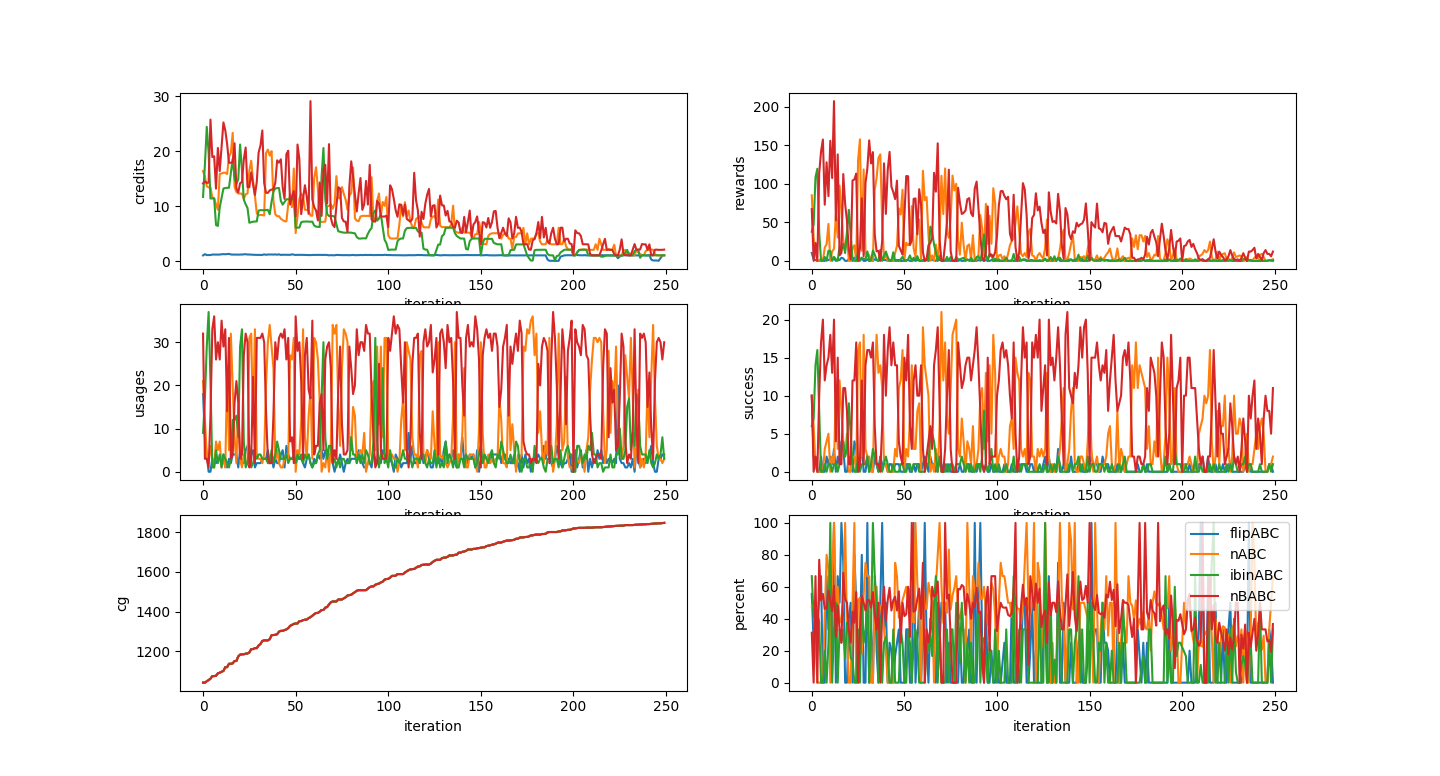}
    \caption{The operator information graphs of the \textit{OneMax} problem}
    \label{fig:OF_RunII_OM}
\end{figure}

Figure \ref{fig:OF_RunII_OM} depicts the performance obtained by the four operators through iterations on a OneMax problem instance with $4000$ dimensions over a single run. The figure consists of $6$ sub-figures; \textit{top-left} indicates the progression of credit level, \textit{top-right} presents the change in rewards, \textit{mid-left} shows the changes in usage of the operators, \textit{mid-right} displays the changes in the success of the operators, \textit{bottom-left }shows the overall convergence, while \textit{bottom-right} figure presents the selectibility of the operators in percentile. All the sub-figures demonstrate the approximation of the behaviours over increasing iterations. This suggests the progression of learning on how to select the operators subject to the circumstances. The search process is split into five regions. It is clearly shown that \textit{n}ABC and \textit{nb}ABC are mostly used operators during the search process. Both received greater rewards than the others, most likely as a result of the update mechanisms of each being better suited to address the \textit{OneMax} problems. It should be noted that \textit{flip}ABC can change only one bit per operation, hence the quality of the solution can maximally increase by $1$ only. As a result, this operator is less frequently utilised.

Table \ref{tab:sukp_2} depicts the comparative results of the four variants of RL-based ABC in terms of 'Rank', 'Max', 'Mean', 'Std', and additionally the 'p-values' from the statistical test to solving the \textit{SUKP} benchmark instances. The overall mean ranks of the variants of \textit{One-Run w-L}, \textit{One-Run w/L}, \textit{All-Run w-L}, and \textit{All-Run w/L} are $2.17$, $2.43$, $2.97$ and $2.43$, respectively, where the best mean rank is $2.17$ achieved by the \textit{One-Run w-L}. \textit{One-Run w/L} and \textit{All-Run w/L} -- the variants with transfer learning -- produce with an equal performance level with the score of $2.43$ in mean rank, which remain slightly worse. In terms of 'Max' column, both \textit{One-Run w-L} and \textit{All-Run w/L} have achieved the best solutions; $14$ times out of the $30$ instances, while \textit{One-Run w/L} and \textit{All-Run w-L} have produced $13$ and $11$ times best out of $30$ instances, respectively.

It is observed that, transfer learning appears to help the \textit{All-Run} variant while having little effect on the 
 \textit{One-Run} variant. Although the work by \cite{durgut2022transfer} demonstrated the benefit of transfer learning with preliminary result and the performance of both \textit{One-Run} and \textit{All-Run} variants over solving \textit{OneMax} problems supporting this idea, further research and investigation would be necessary to see the real benefit of transfer learning in this regard. It appears that the varied qualities of the implemented operators used to solve the two problems, which differ in structure and properties, contribute to the success of transfer learning in \textit{OneMax} but not in the \textit{SUKP} problems. The issue may be a result of the binary operators being employed being more complementary to one another for one problem type but not for the other. For instance, \textit{flip}ABC operator changes only one bit each time once it has been selected, which may help one problem to rescue from local optima, but not for the other. This leads to the possibility that the operators promoted by learning for \textit{OneMax} might not be as effective for \textit{SUKP}. Obviously, this suggests  that further research and investigations are necessary on the underlying RL configuration and the selection of operators in the pool. 


\begin{figure}
    \centering
    \includegraphics[width=\textwidth]{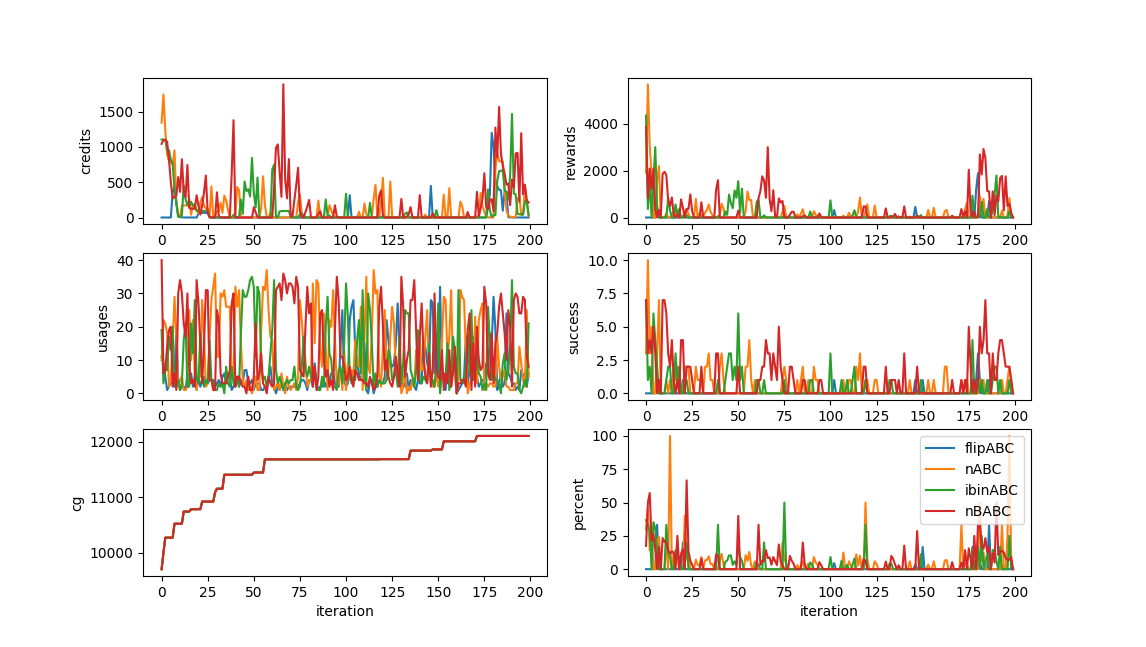}
    \caption{The operator information graph of the \textit{SUKP} problem}
    \label{fig:OF_RUN4_SUKP}
\end{figure}

Figure~\ref{fig:OF_RUN4_SUKP} represents the convergence behaviour of the variants of \textit{SUKP} problem. This figure has also been plotted in the same way as Figure \ref{fig:OF_RunII_OM}, where $6$ sub-figures are included indicating the same set of measures. It is observed that \textit{All-Run w/L} and \textit{One-Run w-L} are the fastest variants on convergence to the optimal solution. Also, the figure shows the performances gained by the four operators through iterations on \textit{SUKP} problem. It is observed that the operator selection is very competitive during the search process at the beginning, (i.e., in the early stages) of the search process in which \textit{n}ABC and \textit{nb}ABC are more dominant, while in the later stages all operators participate in the selection process. A stable non-progressive situation is observed between iterations $75$ and $175$, which can be  clearly seen from all the sub-graphs. The operators are more proactive in the rest of the process over the iterations. This might be due to a special circumstance with the benchmark instances used for this single run.  Although the operators did not achieve success between iterations $75$ and $175$, they eventually turned to escape from local optima, most likely using the \textit{flip}ABC and \textit{n}BABC operators, which appear to be more effective than the other two.

 
\begin{figure}
    \centering
    \includegraphics[width=0.8\textwidth]{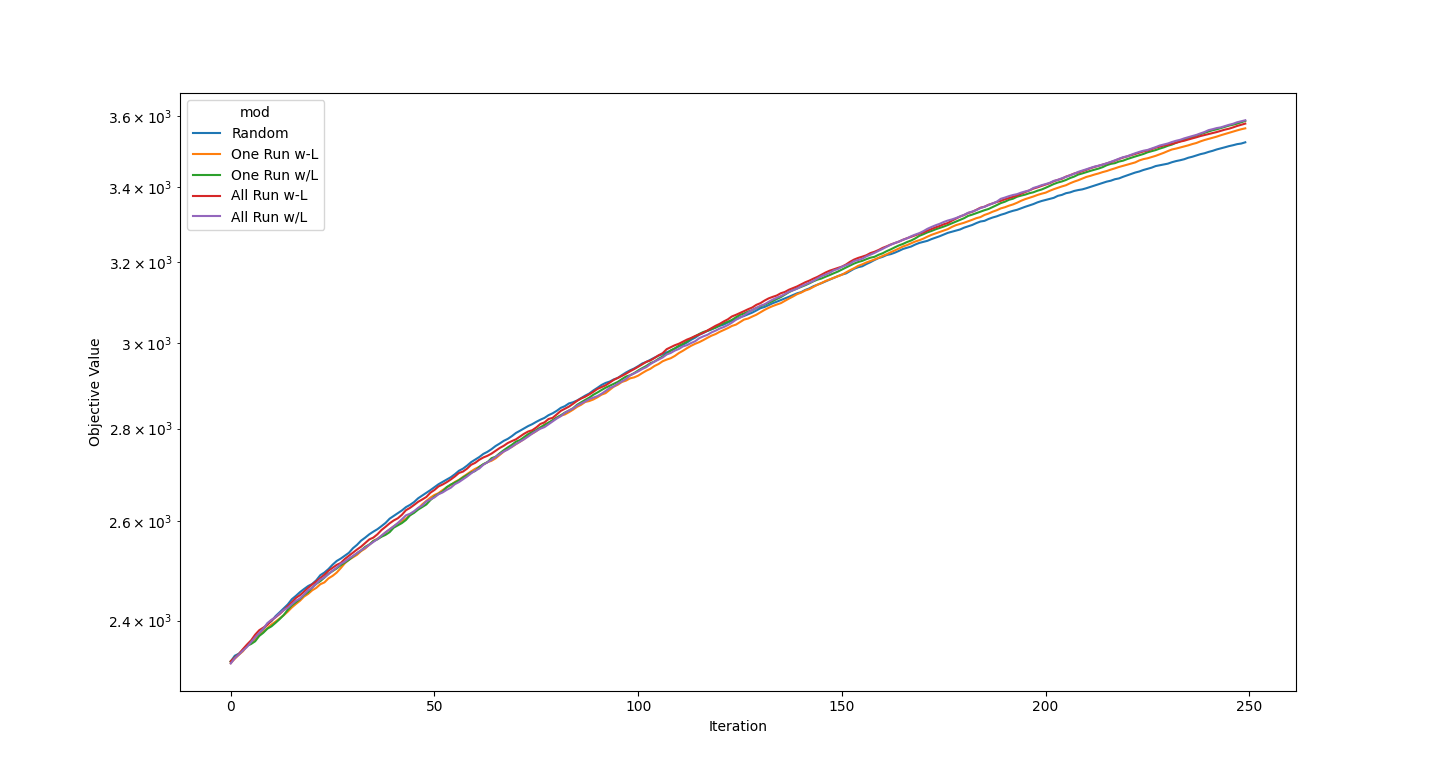}
    \caption{The convergence graph of the  \textit{OneMax} problem}
    \label{fig:CG_1_OM}
\end{figure}
\begin{figure}
    \centering
    \includegraphics[width=0.8\textwidth]{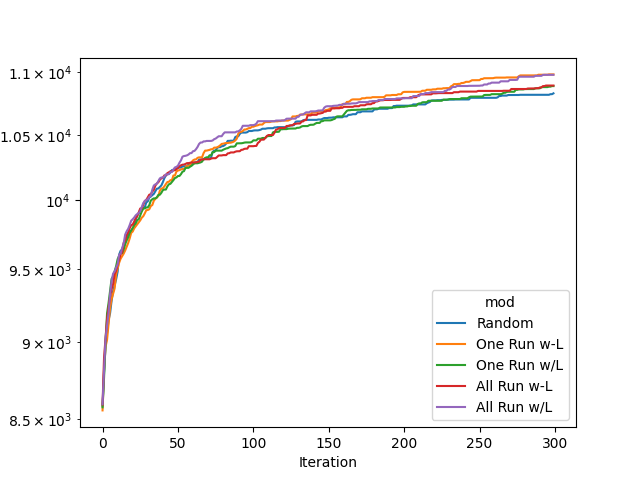}
    \caption{The convergence graph of the  \textit{SUKP} problem}
    \label{fig:CG_5_SUKP}
\end{figure}

The overall convergence achieved by all the variants of the algorithm for solving both \textit{OneMax} and \textit{SUKP} problems are plotted in Figure \ref{fig:CG_1_OM} and \ref{fig:CG_5_SUKP}. It can be observed that, the plots in both cases initially appear to be highly overlapping, but after $175$ iterations in the case of \textit{OneMax} and after $40$ in the case of \textit{SKUP}, they begin to diverge. The overall performance depicted appear to be very consistent with the tabulated results. Also, \textit{OneMax} problems are not much complicated compared to the \textit{SUKP} problems. 
 
\begin{table}
\centering
\caption{Comparative results using the variant algorithms under investigation for the \textit{OneMax} problem instances}\label{tab:OneMax_2}
\smallskip\noindent
\resizebox{1.0\linewidth}{!}{%
\begin{tabular}{ccrrr|crrr|r||crrr|crrr|r}
& \multicolumn{4}{c}{\textit{\textbf{One-Run w-L}}} & \multicolumn{5}{c}{\textit{\textbf{One-Run w/L}}}    & \multicolumn{4}{c}{\textit{\textbf{All-Run w-L}}}    & \multicolumn{5}{c}{\textit{\textbf{All-Run w/L}}}    \\ 
\hline
\multicolumn{1}{l}{ID} & \multicolumn{1}{c}{Rank} & \multicolumn{1}{c}{Max} & \multicolumn{1}{c}{Mean} & \multicolumn{1}{c|}{Std} & \multicolumn{1}{c}{Rank} & \multicolumn{1}{c}{Max} & \multicolumn{1}{c}{Mean} & \multicolumn{1}{c}{Std} & \multicolumn{1}{|c||}{p} & \multicolumn{1}{c}{Rank} & \multicolumn{1}{c}{Max} & \multicolumn{1}{c}{Mean} & \multicolumn{1}{c|}{Std} & \multicolumn{1}{c}{Rank} & \multicolumn{1}{c}{Max} & \multicolumn{1}{c}{Mean} & \multicolumn{1}{c}{Std} & \multicolumn{1}{|c}{p}  \\ 
\hline
 1 & 3  &  500  &  499.50  &  0.81  & 2  &  500   &  499.53  &  0.67  & 9.1E-01  & 1  &  500  &  499.80  &  0.40  & 4  &  500  &  499.27  &  0.81  & 1.1E-02   \\
 2 & 1  &  749  &  744.77  &  2.29  & 2  &  749   &  744.70  &  3.08  & 7.2E-01  & 3  &  748  &  744.67  &  2.10  & 4  &  749  &  743.63  &  2.61  & 7.7E-02   \\
 3 & 2  &  989  &  984.03  &  3.38  & 1  &  989   &  984.87  &  2.78  & 2.6E-01  & 3  &  990  &  983.60  &  3.48  & 4  &  992  &  981.40  &  4.22  & 4.0E-02   \\
 4 & 3  & 1224  & 1214.87  &  5.73  & 1  & 1228   & 1216.77  &  5.49  & 1.5E-01  & 2  & 1224  & 1216.57  &  4.99  & 4  & 1223  & 1209.17  &  6.16  & 3.7E-04   \\
 5 & 3  & 1448  & 1437.20  &  7.18  & 2  & 1453   & 1437.70  &  7.52  & 9.6E-01  & 1  & 1454  & 1438.60  &  8.65  & 4  & 1442  & 1430.20  &  6.81  & 2.3E-04   \\
 6 & 3  & 1664  & 1646.37  &  8.62  & 1  & 1663   & 1649.20  &  7.57  & 1.0E-01  & 2  & 1668  & 1646.60  &  9.61  & 4  & 1657  & 1645.17  &  6.46  & 5.4E-01   \\
 7 & 3  & 1861  & 1841.97  & 13.73  & 2  & 1862   & 1843.97  &  9.83  & 9.4E-01  & 4  & 1873  & 1841.90  & 13.34  & 1  & 1880  & 1856.00  & 11.75  & 3.6E-04   \\
 8 & 4  & 2054  & 2030.77  & 14.57  & 2  & 2066   & 2037.60  & 13.22  & 4.7E-02  & 3  & 2058  & 2031.90  & 15.81  & 1  & 2079  & 2057.50  & 9.92   & 4.1E-06   \\
 9 & 4  & 2251  & 2211.33  & 19.74  & 2  & 2258   & 2223.97  & 16.98  & 2.7E-02  & 3  & 2246  & 2218.33  & 14.97  & 1  & 2283  & 2257.03  & 12.16  & 2.6E-06   \\
10 & 3  & 2435  & 2395.33  & 18.65  & 2  & 2449   & 2400.23  & 19.38  & 3.9E-01  & 4  & 2445  & 2395.03  & 19.87  & 1  & 2475  & 2448.07  & 12.56  & 1.7E-06   \\
11 & 2  & 2621  & 2572.20  & 17.42  & 3  & 2601   & 2572.00  & 21.10  & 6.3E-01  & 4  & 2608  & 2563.23  & 19.30  & 1  & 2655  & 2639.13  & 10.48  & 1.7E-06   \\
12 & 3  & 2786  & 2747.77  & 18.98  & 2  & 2790   & 2752.03  & 19.54  & 3.5E-01  & 4  & 2801  & 2743.90  & 23.20  & 1  & 2850  & 2819.70  & 15.75  & 2.1E-06   \\
13 & 3  & 2937  & 2903.00  & 18.34  & 2  & 2954   & 2923.27  & 23.73  & 6.9E-04  & 4  & 2943  & 2902.27  & 20.55  & 1  & 3026  & 3001.80  & 13.21  & 1.7E-06   \\
14 & 4  & 3125  & 3081.50  & 20.47  & 2  & 3151   & 3092.20  & 24.86  & 1.9E-01  & 3  & 3157  & 3083.47  & 29.91  & 1  & 3201  & 3177.63  & 15.88  & 1.7E-06   \\
15 & 4  & 3305  & 3250.47  & 24.49  & 2  & 3321   & 3263.50  & 28.86  & 3.7E-02  & 3  & 3293  & 3251.60  & 29.76  & 1  & 3383  & 3353.53  & 16.52  & 1.7E-06   \\
16 & 4  & 3462  & 3404.60  & 25.20  & 2  & 3487   & 3421.37  & 36.35  & 1.3E-02  & 3  & 3491  & 3418.50  & 31.27  & 1  & 3573  & 3531.77  & 17.03  & 1.7E-06   \\
17 & 4  & 3642  & 3563.90  & 34.78  & 2  & 3647   & 3585.27  & 39.62  & 2.1E-02  & 3  & 3621  & 3577.47  & 24.08  & 1  & 3761  & 3707.33  & 19.51  & 1.7E-06   \\
18 & 4  & 3797  & 3739.50  & 28.98  & 2  & 3797   & 3750.53  & 33.00  & 2.7E-01  & 3  & 3813  & 3750.00  & 33.75  & 1  & 3918  & 3880.30  & 19.53  & 1.7E-06   \\
19 & 4  & 3951  & 3904.90  & 34.57  & 3  & 3980   & 3909.50  & 41.29  & 6.8E-01  & 2  & 3965  & 3911.13  & 25.10  & 1  & 4078  & 4043.13  & 16.73  & 1.7E-06   \\ 
\hline
& 3.21   &     &    &   & 1.94   &   &    &   &   & 2.89  &    &    &  & 1.94  &    &    &   & \\ \hline  
\end{tabular}}
\end{table}

\begin{table}
\centering
\caption{Comparative results using the variant algorithms under investigation for the \textit{SUKP} problem instances}
\label{tab:sukp_2}
\smallskip\noindent
\resizebox{1.0\linewidth}{!}{%
\begin{tabular}{lcrrr|crrr|r||crrr|crrr|r}
& \multicolumn{4}{c}{\textit{\textbf{One-Run w-L}}}  & \multicolumn{5}{c}{\textit{\textbf{One-Run w/L}}} & \multicolumn{4}{c}{\textit{\textbf{All-Run}} w-L}   & \multicolumn{5}{c}{\textit{\textbf{All-Run w/L}}}      \\ 
\cline{1-19}
ID    & \multicolumn{1}{l}{Rank} & \multicolumn{1}{c}{Max} & \multicolumn{1}{c}{Mean} & \multicolumn{1}{c}{Std} & \multicolumn{1}{|c}{Rank} & \multicolumn{1}{c}{Max} & \multicolumn{1}{c}{Mean} & \multicolumn{1}{c}{Std} & \multicolumn{1}{|c||}{p} & \multicolumn{1}{c}{Rank} & \multicolumn{1}{c}{Max} & \multicolumn{1}{c}{Mean} & \multicolumn{1}{c}{Std} & \multicolumn{1}{|c}{Rank} & \multicolumn{1}{c}{Max} & \multicolumn{1}{c}{Mean} & \multicolumn{1}{c}{Std} & \multicolumn{1}{|c}{p}  \\ 
\cline{1-19}
1\_1  & 1  & 13167  & 13052.20  & 30.68  & 2 & 13167  & 13046.23  & 23.49   & 3.6E-01  & 4  & 13044  & 13034.87  & 44.27   & 3  & 13046 & 13042.20  & 7.01 & 6.8E-01  \\
1\_2  & 1  & 12274 & 12076.95  & 65.72   & 2 & 12235  & 12023.00  & 122.22  & 1.4E-01  & 3  & 12130  & 11993.63  & 168.86  & 4  & 12274 & 11979.93  & 181.49 & 9.3E-01 \\
1\_3  & 1  & 13340 & 13165.27  & 106.95  & 4 & 13340  & 13107.37  & 117.23  & 1.5E-01  & 3  & 13319  & 13126.80  & 109.97  & 2  & 13402 & 13148.77  & 116.7  & 3.9E-01 \\
1\_4  & 1  & 13671 & 13376.13  & 191.36  & 3 & 13671  & 13319.03  & 210.72  & 2.5E-01  & 4  & 13671  & 13297.73  & 205.28  & 2  & 13660 & 13319.70  & 177.16 & 7.0E-01  \\
1\_5  & 1  & 10980 & 10588.03  & 160.02  & 4 & 10968  & 10537.80  & 152.81  & 1.4E-01  & 3  & 10896  & 10548.63  & 132.44  & 2  & 10972 & 10549.00  & 200.13 & 9.3E-01  \\
1\_6  & 4  & 12012 & 11355.87  & 308.24  & 1 & 12135  & 11502.87  & 318.34  & 4.3E-02  & 3  & 12035  & 11383.47  & 278.75  & 2  & 12135 & 11444.43  & 330.27 & 3.8E-01  \\
1\_7  & 1  & 10994 & 10616.67  & 180.01  & 2 & 11001  & 10594.87  & 216.73  & 6.6E-01  & 3  & 11244  & 10580.90  & 231.11  & 4  & 10944 & 10537.27  & 206.51 & 5.2E-01  \\
1\_8  & 1  & 10168 & 10055.10  & 137.99  & 4 & 10168  & 10032.90  & 168.48  & 5.2E-01  & 2  & 10182  & 10036.23  & 139.19  & 3  & 10175 & 10034.13  & 149.66 & 8.7E-01  \\
1\_9  & 2  & 11426 & 11118.60  & 150.97  & 1 & 11296  & 11120.30  & 129.08  & 9.7E-01  & 4  & 11296  & 11042.73  & 127.75  & 3  & 11296 & 11054.93  & 130.83 & 9.8E-01  \\
1\_10 & 1  &  9409 &  9166.57  & 127.46  & 4  & 9486  & 9105.27   & 138.28  & 4.8E-02  & 3  &  9486  &  9140.33  & 159.16  & 2  &  9494 &  9144.17  & 201.14 & 9.8E-01  \\
2\_1  & 3  & 14044 & 13850.57 & 79.48  & 1  & 13963  & 13852.63  & 59.98  & 8.8E-01   & 4  & 13963  & 13843.27 & 68.49  & 2  & 14044  & 13850.63  & 75.75  & 7.6E-01   \\
2\_2  & 4  & 13407 & 13222.67 & 149.27  & 1  & 13498  & 13297.03  & 162.32 & 8.9E-02  & 2  & 13407  & 13278.13  & 120.32  & 3  & 13498  & 13257.77 & 179.72 & 4.7E-01  \\
2\_3  & 1  & 12350 & 11792.00 & 253.90  & 3  & 12328  & 11774.40  & 230.48  & 4.7E-01  & 4  & 12242  & 11708.17  & 235.69  & 2  & 12103  & 11780.77 & 228.71 & 3.7E-01  \\
2\_4  & 3  & 11821 & 11550.13 & 257.17  & 2  & 11846  & 11592.20  & 211.66  & 4.7E-01  & 1  & 11830  & 11613.43  & 192.09  & 4  & 12187  & 11545.47 & 252.33 & 2.5E-01  \\
2\_5  & 1  & 12644 & 12523.10 & 166.57  & 3  & 12644  & 12462.20  & 167.07  & 1.4E-01  & 4  & 12644  & 12417.17  & 228.26  & 2  & 12644  & 12484.97 & 187.49 & 2.2E-01 \\
2\_6  & 1  & 10735 & 10618.93 & 118.84  & 4  & 10666  & 10558.13  & 201.66  & 1.9E-01  & 2  & 10731  & 10571.00  & 186.18  & 3  & 10666  & 10569.10 & 157.88 & 8.8E-01  \\
2\_7  & 3  & 11001 & 10832.70 & 112.07  & 1  & 11282  & 10870.67  & 152.82  & 3.4E-01  & 4  & 11160  & 10831.37  & 135.25  & 2  & 11110  & 10834.43 & 141.60 & 9.4E-01  \\
2\_8  & 3  & 10089 & 9665.67  & 170.46  & 1  & 10355  & 9720.03   & 222.45  & 3.4E-01  & 4  & 10250  & 9600.00  & 208.11  & 2  & 10355  & 9666.77  & 213.64 & 4.5E-01  \\
2\_9  & 2  & 10734 & 10592.90 & 94.68  & 3  & 10735 & 10587.90  & 88.78  & 9.8E-01  & 4  & 10784  & 10585.33  & 94.82  & 1  & 10902 & 10603.00  & 100.29 & 5.3E-01   \\
2\_10 & 2  & 10176 & 9651.73   & 219.47  & 1  & 10176  & 9720.33  & 199.58  & 9.6E-01 & 3  & 10176  & 9592.83  & 172.32  & 4 & 9977  & 9588.27 & 188.11  & 5.4E-01   \\
3\_1  & 4  & 11579 & 11348.37  & 96.03  & 2  & 12020  & 11392.00  & 181.24  & 3.8E-01 & 1  & 12045  & 11409.47  & 201.67 & 3  & 11752 & 11386.73  & 120.57 & 7.9E-01  \\
3\_2  & 4  & 12369  & 11864.07 & 289.56  & 3  & 12369  & 11917.87  & 282.67  & 4.7E-01 & 2  & 12369  & 11923.83  & 345.64  & 1 & 12369  & 11950.50 & 288.43 & 7.5E-01  \\
3\_3  & 1  & 13392 & 13141.20  & 174.84  & 4  & 13500  & 13118.67  & 185.12  & 7.8E-01 & 3 & 13477  & 13119.73  & 211.00 & 2  & 13458  & 13132.27 & 225.99  & 7.5E-01  \\
3\_4  & 4  & 10920 & 10620.63  & 138.02  & 1  & 10920  & 10675.17  & 161.28  & 2.3E-01  & 3  & 10920  & 10623.07 & 118.08  & 2 & 10920  & 10642.53 & 136.11 & 6.1E-01  \\
3\_5  & 1  & 11538  & 11154.93 & 147.54  & 3  & 11538  & 11099.40  & 129.45  & 1.4E-01  & 2  & 11538  & 11101.37 & 172.69  & 4 & 11538  & 11085.17 & 137.35 & 8.4E-01  \\
3\_6  & 4  & 11414 & 10941.50 & 218.08 & 2 & 11377 & 10953.23 & 213.23  & 6.4E-01 & 3  & 11237  & 10944.23  & 187.32  & 1  & 11377  & 10980.60 & 235.73  & 4.7E-01   \\
3\_7  & 2  & 10068  & 9911.37  & 70.59  & 3 & 10026  & 9898.70  & 64.92 & 4.5E-01 & 4  & 10294  & 9898.60 & 100.12 & 1 & 10176   & 9916.87    & 82.33 & 2.5E-01  \\
3\_8  & 2  &  9588 & 9294.77  & 138.33   & 3   & 9496  & 9294.43   & 119.37   & 9.6E-01  & 4  & 9532   & 9262.50  & 139.24  & 1  & 9796  & 9297.50   & 177.36  & 5.4E-01 \\
3\_9  & 3  & 10653  & 10494.07  & 84.91  & 1  & 10636  & 10511.57  & 73.70  & 3.8E-01  & 2  & 10732  & 10504.37  & 97.57  & 4  & 10636  & 10482.23  & 95.90  & 5.6E-01  \\
3\_10 & 3  &  9514  & 9359.17  & 142.50  & 4  & 9561   & 9299.63   & 144.89   & 1.2E-01  & 1  & 9514   & 9380.23  & 136.25   & 2   & 9514  & 9366.73  & 88.95  & 4.0E-01 \\
\cline{1-19}
 Mean     & 2.17   &     &    &   & 2.43    &   &     &    &   & 2.97     &     &    &    & 2.43    &    &    &    & \\ \hline 
\end{tabular}}
\end{table}




\begin{table}
\caption{Comparative results considering the state-of-the-art works for the \textit{SUKP} problem}
\label{tab:sukp_3}
\scriptsize
\centering
\resizebox{1.0\linewidth}{!}{
\begin{tblr}{
  cell{1}{2,4,6,8,10,12,14} = {c=2}{c},
  cell{3-33}{1-15} = {c=1}{r},
  hline{2-3,33} = {-}{},
}\hline
      & A-SUKP &       & GA    &       & BABC  &       & ABCBin &       & binDE &       & GPSO* &       & Proposed &          \\
ID    & Max    & Mean  & Max   & Mean  & Max   & Mean  & Max    & Mean  & Max   & Mean  & Max   & Mean  & Max      & Mean     \\
1\_1  & 12459  & 12459 & 13044 & 12956 & 13251 & 13029 & 13044  & 12819 & 13044 & 12991 & 13167 & 12937 & 13167    & 13052.20  \\
1\_2  & 11119  & 11119 & 12066 & 11546 & 12238 & 12155 & 12238  & 12049 & 12274 & 12124 & 12210 & 11778 & 12274    & 12076.95 \\
1\_3  & 11292  & 11292 & 13064 & 12493 & 13241 & 13064 & 12946  & 11862 & 13241 & 12941 & 13302 & 12766 & 13340    & 13165.27 \\
1\_4  & 12262  & 12262 & 13671 & 12803 & 13829 & 13359 & 13671  & 12537 & 13671 & 13110 & 13993 & 12949 & 13671    & 13376.13 \\
1\_5  & 8941   & 8941  & 10553 & 9981  & 10428 & 9995  & 9751   & 9339  & 10420 & 9899  & 10600 & 10090 & 10980    & 10588.03 \\
1\_6  & 9432   & 9432  & 11016 & 10350 & 12012 & 10903 & 10913  & 9958  & 11661 & 10499 & 11935 & 10750 & 12012    & 11355.87 \\
1\_7  & 9076   & 9076  & 10083 & 9642  & 10766 & 10065 & 9674   & 9188  & 10576 & 9681  & 10698 & 9947  & 10994    & 10616.67 \\
1\_8  & 8514   & 8514  & 9831  & 9327  & 9649  & 9136  & 8978   & 8540  & 9649  & 9021  & 10168 & 9417  & 10168    & 10055.10  \\
1\_9  & 9864   & 9864  & 11031 & 10568 & 10784 & 10452 & 10340  & 9910  & 10586 & 10364 & 11258 & 10566 & 11426    & 11118.60  \\
1\_10 & 8299   & 8299  & 9472  & 8693  & 9090  & 8858  & 8789   & 8364  & 9191  & 8784  & 9759  & 8779  & 9409     & 9166.57  \\
2\_1  & 13634  & 13634 & 14044 & 13806 & 13860 & 13735 & 13860  & 13547 & 13814 & 13676 & 13963 & 13740 & 14044    & 13850.57 \\
2\_2  & 11325  & 11325 & 13145 & 12235 & 13508 & 13352 & 13498  & 13103 & 13407 & 13212 & 13498 & 12937 & 13407    & 13222.67 \\
2\_3  & 10328  & 10328 & 11656 & 10889 & 11846 & 11194 & 11191  & 10424 & 11535 & 10969 & 11972 & 11233 & 12350    & 11792.00    \\
2\_4  & 9784   & 9784  & 11792 & 10828 & 11521 & 10945 & 11287  & 10346 & 11469 & 10717 & 12167 & 11027 & 11821    & 11550.13 \\
2\_5  & 10208  & 10208 & 12055 & 11755 & 12186 & 11946 & 11494  & 10922 & 12304 & 11865 & 12736 & 11934 & 12644    & 12523.10  \\
2\_6  & 9183   & 9183  & 10666 & 10099 & 10382 & 9860  & 9633   & 9187  & 10382 & 9710  & 10724 & 9907  & 10735    & 10618.93 \\
2\_7  & 9751   & 9751  & 10570 & 10112 & 10626 & 10101 & 10160  & 9549  & 10462 & 9976  & 11048 & 10400 & 11001    & 10832.70  \\
2\_8  & 8497   & 8497  & 9235  & 8794  & 9541  & 9033  & 9033   & 8366  & 9388  & 8768  & 10264 & 9195  & 10089    & 9665.67  \\
2\_9  & 9615   & 9615  & 10460 & 10185 & 10755 & 10328 & 10071  & 9738  & 10546 & 10228 & 10647 & 10205 & 10734    & 10592.90  \\
2\_10 & 7883   & 7883  & 9496  & 8883  & 9318  & 9181  & 9262   & 9618  & 9312  & 9096  & 9839  & 9107  & 10176    & 9651.73  \\
3\_1  & 10231  & 10231 & 11454 & 11093 & 11664 & 11183 & 11206  & 10880 & 11352 & 11075 & 11710 & 11237 & 11579    & 11348.37 \\
3\_2  & 10483  & 10483 & 12124 & 11326 & 12369 & 12082 & 12006  & 11485 & 12369 & 11876 & 12369 & 11684 & 12369    & 11864.07 \\
3\_3  & 11508  & 11508 & 12841 & 12237 & 13047 & 12523 & 12308  & 11668 & 13024 & 12278 & 13298 & 12514 & 13392    & 13141.20  \\
3\_4  & 8621   & 8621  & 10920 & 10352 & 10602 & 10151 & 10376  & 9684  & 10547 & 10085 & 10856 & 10208 & 10920    & 10620.63 \\
3\_5  & 9961   & 9961  & 10994 & 10640 & 11158 & 10776 & 10269  & 9957  & 11152 & 10661 & 11310 & 10762 & 11538    & 11154.93 \\
3\_6  & 9618   & 9618  & 11093 & 10190 & 10528 & 9898  & 10051  & 9424  & 10528 & 9832  & 11226 & 10309 & 11414    & 10941.50  \\
3\_7  & 8672   & 8672  & 9799  & 9433  & 10085 & 9538  & 9235   & 8905  & 9883  & 9315  & 8971  & 9552  & 10068    & 9911.37  \\
3\_8  & 8064   & 8064  & 9173  & 8704  & 9456  & 9090  & 8932   & 8407  & 9352  & 8847  & 9389  & 8881  & 9588     & 9294.77  \\
3\_9  & 9559   & 9559  & 10311 & 9993  & 10823 & 10483 & 10537  & 9615  & 10728 & 10159 & 10595 & 10145 & 10653    & 10494.07 \\
3\_10 & 8157   & 8157  & 9329  & 8849  & 9333  & 9086  & 8799   & 8348  & 99218 & 8920  & 9807  & 8917  & 9514     & 9359.17  \\ 
\end{tblr}
}
\end{table}

\begin{figure}
    \centering
    \includegraphics[width=0.6\linewidth]{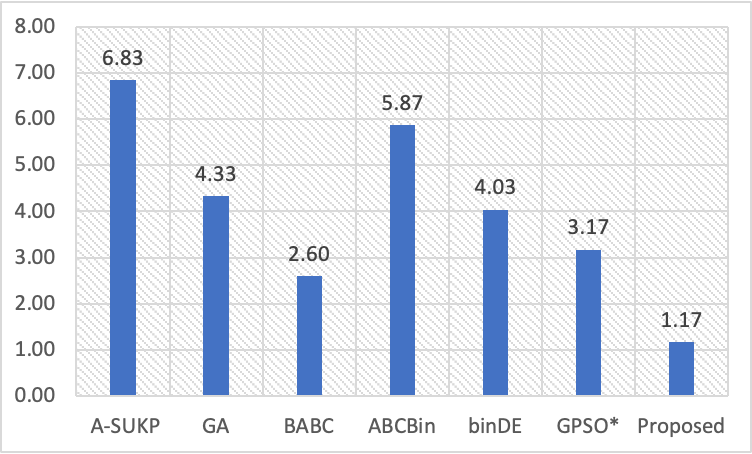}
    \caption{The average ranks calculated over the comparative performances of the state-of-the art approaches to solving the \textit{SUKP} problem instances}
    \label{fig:rankbars}
\end{figure}

In order to further demonstrate the performance of the proposed approach for \textit{SUKP}, a comparison analysis is conducted with the state-of-the-art algorithms. Table \ref{tab:sukp_3} presents the comparative results, where the proposed algorithm outperforms the others in most of the cases. The algorithms considered for comparisons have been selected from the most relevant literature, where A-SUKP, GA, BABC, ABCBin, and binDE results were captured from \cite{he2018novel, wu2020solving} and GPSO* is from \cite{ozsoydan2019swarm}, which are published within the last $5$ years. Among them, A-SUKP is a greedy local search algorithm while the others are population-based metaheuristic algorithms; GA and binDE are evolutionary algorithms, GPSO* is a particle swarm optimisation algorithm, and the rest are the variants of binary ABC algorithms, which are the most relevant approaches. 

Table \ref{tab:sukp_3} tabulates the 'Mean' and 'Max' scores of each approaches for the benchmark instances, while Figure \ref{fig:rankbars} plots the average rankings, which are indicated as $6.83$, $4.33$, $2.60$, $5.87$, $4.03$, $3.17$ and $1.17$ for A-SUKP, GA, BABC, ABCBin, bin DE, GPSO* and 'Proposed', respectively. The proposed approach is the outperforming variant, \textit{One-Run w-L}, from Table \ref{tab:sukp_2}. Figure \ref{fig:rankbars} confirms that the best rank among the compared algorithms clearly goes to 'Proposed' approach with the score of $1.17$ while the first runner is BABC with $2.60$ and the second runner is GPSO* with $3.17$. 

\section{Conclusions and Future Work}\label{concs}
This article presented a comprehensive research and experiment on (i) how to gain experience in search processes in order to adaptively select operators using Reinforcement Learning (RL); (ii) how to develop a scalable approach such that the operator selection agent can be trained once and used in a variety of similar and other circumstances; and (iii) how to transfer the gained experience across the problem instances and the types. 

This research work serves as a proof-of-concept to show how RL could potentially be used to develop adaptive operator selection schemes for the purposes itemised above. The study has demonstrated that items (i) and (ii) are achieved successfully, while item (iii) has been partially accomplished. This requires further research and investigations into more advanced and sophisticated RL models to be considered and applied for the operators selection problem. In addition, the set of binary operators has significance in terms of how well they complement one another. As a highly unknown problem type, the search problem requires further improvements to the learning quality. This study has considered both the traditional RL techniques and the structural credit assignment. The authors are aware of how deep learning models can be utilised to increase the effectiveness of RL by precisely mapping states to actions and by producing better predictions. Moreover, the notion of multi-agent reinforcement learning and collective intelligence techniques can further be developed and integrated into this study for the same purpose. 


\section{Acknowledgements}
The numerical calculations reported in this paper were fully/partially performed at TUBITAK ULAKBIM, High Performance and Grid Computing Center (TRUBA resources).
The second author was supported by the TUBITAK-2219 program.

\appendix
\section{Feature functions}\label{sec:appendix}
The list of equations used to calculate the features that characterise the problem states, where Eq:~\ref{eq:phi_1} to \ref{eq:phi_11} are related to the population and the rest are related to the individual solutions. The base notation of population-based features is as follows. Let $P=\{p_i|i=0,1,...,N\}$ be the set of parent solutions and $C=\{c_i|i=0,1,...,N\}$ be the set of children solutions reproduced from $P$, where each solution has $D$ dimensions. Also, let $F^p=\{f^p_i|i=0,1,...,N\}$ be the set of parent fitness values and $F^c=\{f^c_i|i=0,1,...,N\}$ be set of children fitness values. $g_{best}$ represents the best solution that has been found so far and $p_{best}$ represents the best solution in the current population. 

\begin{eqnarray}
    \phi_1 &=& \frac{\sum^{n-1}_{i=0}{\sum^n_{j=i+1}{\|p_i-p_j\|}}}{D \frac{n(n-1)}{2}} \label{eq:phi_1}\\ [8pt]
    \phi_2 &=& \frac{\sum^{n-1}_{i=0}{\sum^n_{j=i+1}{\|f^p_i-f^p_j\|}}}{\frac{n(n-1)}{2}}\label{eq:phi_2}\\ [8pt]
    \phi_3 &=&   \begin{cases}
                            \frac{|c_i|}{N}       & \quad \text{if } f^c_i > f^p_i \\
                                0  & \quad \text{otherwise }
                        \end{cases} \label{eq:phi_3}\\ [8pt] 
    \phi_4 &=&  \begin{cases}
                            \frac{|c_i|}{N}       & \quad \text{if } f^c_i > g_{best} \\
                                0  & \quad \text{otherwise }
                        \end{cases} \label{eq:phi_4} \\ [8pt]
    \phi_5 &=&  \begin{cases}
                            \frac{\frac{f^c_i - f^p_i}{f^c_i}}{N}       & \quad \text{if } f^c_i > f^p_i \\
                                0  & \quad \text{otherwise }
                        \end{cases} \label{eq:phi_5} \\ [8pt] 
\end{eqnarray}

\begin{eqnarray}
    \phi_6 &=&  \frac{E[\max(F^c) - \max(F^p)]}{\max(F^p)} \label{eq:phi_6}\\[8pt] 
    \phi_7 &=&  \frac{E[\|x^* - x_t\| - \|x^* - x_{t+1}\|]}{D}\label{eq:phi_7}\\[8pt]     
    \phi_8 &=&  \begin{cases}
        \sum_{P_{i,j} \in N(P_i)} \frac{\frac{|f^*(P_i) - f(P_{i,j})|}{N} }{\sigma(f(P_i))}  &  N(P_i)\geq 1\\
        0 & \text{otherwise}
        \end{cases}\label{eq:phi_8}\\[10pt] 
    \phi_9 &=&  \phi_4 \times \phi_8 \label{eq:phi_9}\\ [8pt] 
    \phi_{10} &=&   \frac{\sum^{n}_{i}tn_i}{N}\label{eq:phi_10}\\[8pt] 
    \phi_{11} &=&   \max_{i,j \in \{P,C\}}{ \|p_i-c_i\|} \label{eq:phi_11}
\end{eqnarray}

\begin{eqnarray}
    \phi_{12} &=&   \frac{\|x^* - p_i\|}{D} \label{eq:phi_12}\\ [8pt] 
    \phi_{13} &=&   \frac{\|p_i - c_i\|}{D} \label{eq:phi_13}\\ [8pt] 
    \phi_{14} &=&   \frac{f^{x^*} - f^c_i}{f^{x^*}} \label{eq:phi_14}\\ [8pt] 
    \phi_{15} &=&   \frac{f^c_i - f^p_i}{f^c_i}  \label{eq:phi_15}\\ [8pt] 
    \phi_{16} &=&   \frac{\|p_{best} - p_i\|}{D}  \label{eq:phi_16}\\ [8pt] 
    \phi_{17} &=&   \frac{\|p_{worst} - p_i\|}{D}  \label{eq:phi_17}\\ [8pt] 
    \phi_{18} &=&   \frac{\textrm{trail}_i}{\textrm{trial}_{max}}\label{eq:phi_18}\\ [8pt] 
    \phi_{19} &=&   \frac{osi}{csi} \label{eq:phi_19}
\end{eqnarray}

\pagebreak
\section{ABC Algorithm}\label{sec:algorithm}
\begin{algorithm}[h!]
\caption{Artificial Bee Colony}
\label{alg:ABC}
        \begin{algorithmic}[1]
        \State{Initialisation of ABC}
        \While{Termination criteria is not met}
        \State{\textbf{Employed Bee Phase}}
        \For{each bee in colony}
            \State{Produce and evaluate candidate solution} \Comment{operator selected}
            \If{Candidate is better than current bee}
                \State{Replace current bee with candidate}
            \Else
            \State{Increment trial number of current bee}
            \EndIf
        \EndFor
        \State{\textbf{Onlooker Bee Phase}}
        \State{Calculate selection probability of each bee, $p(bee_i)$}
        \For{$i \gets 1$ to $n$}             \Comment{$n$ is the \# of onlooker bees}
            \State{Choose current bee according to $p(bee_i)$}
            \State{Produce and evaluate candidate solution} \Comment{operator selected}
            \If{Candidate is better than current bee}
                \State{Replace current bee with candidate}
            \Else
            \State{Increment trial number of current bee}
            \EndIf
        \EndFor
        \State{\textbf{Scout Bee Phase}}
        \If{trial number of any bee has exceeds the limit}
            \State{Replace the first scout bee with random valid solution}
        \EndIf
        \State{Update the best solution}
        \EndWhile
        \end{algorithmic}
\end{algorithm}

\bibliographystyle{unsrt}  
\bibliography{ref,cas-refs}

\end{document}